\documentclass[review]{elsarticle}
\usepackage{hyperref}
\usepackage{lineno}
\modulolinenumbers[1]
\usepackage{setspace}
\usepackage{amsmath}
\usepackage{amsfonts}
\usepackage{soul}
\usepackage{wrapfig}
\usepackage{color}
\usepackage{subfigure} 
\journal{Engineering Applications of Artificial Intelligence}
\usepackage[rmargin=1.5in, lmargin=1.5in, top=1.5in, bmargin=1.3in]{geometry}

\usepackage{etoolbox}
\makeatletter
\patchcmd{\ps@pprintTitle}% <cmd>
{\let\@oddhead\@empty}% <search>
{\def\@oddhead{\mbox{}\hfill\thepage}}% <replace>
{}{}% <success><failure>
\makeatother
\begin{document}
\begin{frontmatter}

\title{Proportional integral derivative booster for neural networks-based time-series prediction: Case of water demand prediction}
\author[mymainaddress,mysecondaryaddress]{Tony Salloom}
%\ead{Tonysalloom@ieee.org}

\author[mymainaddress,mysecondaryaddress,Mythirdaddres]{Okyay Kaynak}
%\ead{Okyay.kaynak@boun.edu.tr}

\author[mysecondaryaddress]{Xinbo Yu}

\author[mymainaddress,mysecondaryaddress]{Wei He\corref{mycorrespondingauthor}}
\cortext[mycorrespondingauthor]{Corresponding author}
\ead{Weihe@ieee.org}

\address[mymainaddress]{School of Automation and Electrical Engineering, University of Science and Technology Beijing, Beijing 100083, China}
\address[mysecondaryaddress]{Institute of Artificial Intelligence, University of Science and Technology Beijing, Beijing 100083, China}
\address[Mythirdaddres]{Bogazici University, Istanbul Turkey}

\begin{abstract}
% \begin{linenumbers}
Multi-step time-series prediction is an essential supportive step for decision-makers in several industrial areas. Artificial intelligence techniques, which use a neural network component in various forms, have recently frequently been used to accomplish this step. However, the complexity of the neural network structure still stands up as a critical problem against prediction accuracy. In this paper, a method inspired by the proportional-integral-derivative (PID) control approach is investigated to enhance the performance of neural network models used for multi-step ahead prediction of periodic time-series information while maintaining a negligible impact on the complexity of the system. The PID-based method is applied to the predicted value at each time step to bring that value closer to the real value. The water demand forecasting problem is considered as a case study, where two deep neural network models from the literature are used to prove the effectiveness of the proposed boosting method. Furthermore, to prove the applicability of this PID-based booster to other types of periodic time-series prediction problems, it is applied to enhance the accuracy of a neural network model used for multi-step forecasting of hourly energy consumption. The comparison between the results of the original prediction models and the results after using the proposed technique demonstrates the superiority of the proposed method in terms of prediction accuracy and system complexity.
% \end{linenumbers}
\end{abstract}
\end{frontmatter}
\newpage
% \linenumbers
\section{Introduction} \label{Intro}
In recent years, the use of artificial intelligence techniques in the form of machine learning (ML) and deep learning (DL) has swept through the vast majority of the research fields that may come to mind. Their superiority over traditional approaches has been confirmed in several comparative research in literature as in \cite{Papacharalampous2019a, Guo2018, Antunes2018}. The basic component in such approaches is a neural network (NN). In numerous studies, the abilities of NNs are exploited in a number of ways. To mention a few, in the field of robotics, researchers use the approximation ability of NNs to compensate for the uncertain information in the dynamics of the robot as in \cite{He9205695,he2020modeling,Sea2020}. In computer vision, NNs are used for objects detection such as drug bills \cite{Ou2020}, fabric defection \cite{Ouyang2019}, etc. The ability of feature extraction is used for classification and prediction, where Barchi et al. \cite{Barchi2021} prove the efficiency of deep convolution NN in source code classification, in \cite{Song2020}, Song et al. use the extreme learning machine for fault detection and classification, and in \cite{Capizzi2020}, Capizzi et al. build a spiking neural network for long-term prediction of biogas production. The deep belief network is used for PM2.5 concentration prediction in Beijing \cite{Xing2020}.

Time series prediction is one of the most illustrious applications of neural networks (NNs), especially after the rise of the memory power provided by long-short term memory (LSTM) and gated recurrent unit (GRU).
This research is concerned with NN models that are meant to forecast periodic time-series information. Periodicity is a natural phenomenon that characterizes many real-life events. Hourly water and power consumption, daily streamflow, hourly average temperature, hourly pollution rate in the city, and rain rate are but a few examples of periodic time-series information. 
Prediction of time-series information enables planning in several fields of economic and industrial activities \cite{Rosenfeld2020}. Particularly, multi-step prediction is commonly met in real-life scenarios, where planners need to predict future observations based on a given sequence of historical information \cite{Du2018}.
Several papers that propose different NN-based methods for forecasting periodic time series data considering both single-step and multi-step scenarios can be found in the literature. For example, Liu et al. \cite{Liu2020} propose a novel NN model for time series forecasting based on dual-stage two-phase model and temporal attention recurrent NN and prove its applicability in the fields of energy, finance, environment and medicine. While Wang et al. \cite{ Wang2020wind} and Safari et al. \cite{Safari2017} predict one step of short-term wind power interval based on GRU and decomposition approaches. In \cite{Khodayar2017}, Khodayar et al. provide a wind speed prediction method based on a rough deep NN. In \cite{Salloom2021}, GRU and $K$-means classification methods are employed for forecasting quarterly water demand. On the other hand, the outputs of two LSTM networks are integrated to forecast the daily water demand in \cite{Du2021}.
	
Two strategies are used for multi-step time series prediction \cite{BenTaieb2012, Bao2014}: (i) The multi-output strategy, or direct strategy as is called in \cite{Cortez2019} and \cite{Soto2018}, where a multi-input-multi-output model is built to predict several time-steps in the future at one go based on a sequence of historical data as input. This strategy is used in \cite{Soto2019}, where a novel approach is introduced to predict stocks based on fuzzy aggregation models with modular NNs. It is also considered in \cite{Soto2014} for multi-step stocks prediction, where impressive research is conducted to predict the Dow Jones stocks price based on ensembles of adaptive neuro-fuzzy inferences systems models. 
(ii) The iterative strategy where the model is designed for one time-step, then it is iterated to predict the next time-step. In this case, the prior time-step is included in the input of the prediction of the next time-step. Here, a problem comes to one's mind when the iterative technique is used for multi-step ahead prediction, where the values of $ n $ consecutive periods need to be predicted while the dataset is yet to be updated. The reliance on the forecasted values to predict the next values poses the problem of error accumulation \cite{Wang2016a}. Thus the error increases as more predicted values are used.
In literature, some papers are available that attempt to solve this problem, where some authors suggest using different models for different time-steps to reduce the dependency on forecasted values \cite{Ambrosio2019}. Salloom et al. \cite{Salloom2021} apply the $K$-means method to find a relationship between the data at the same time step over previous days and reduce the effects of the predicted value in the input of the next step. Ye et al. \cite{Ye2019} suggest a multi-task-learning algorithm that handles forecasting different horizons as different tasks. Although this kind of solution provides a high level of accuracy, the approach requires extremely high computational time to configure systems, and a large memory to save system parameters, which hinders system maintenance and update processes. On the other hand, some solutions for particular problems are proposed, where Sardinha-Lourenco et al. \cite{Sardinha-Lourenco2018} propose a parallel adaptive weighting strategy to enhance the performance of time-series water consumption prediction. bio-inspired algorithms using fuzzy integrator to predict time-series information. Other works  \cite{Li2018b,Chen2018} suggest the use of a support vector machine to enhance the performance of the LSTM-based NN model.  In \cite{Li2019a}, the authors propose a hybrid system consisting of wavelet packet decomposition and Elman Neural Networks for multi-step ahead wind speed forecasting. Although the authors of these works stated that their strategies have a reasonable computational load, they did not consider the complexity of the system.
Thus, the tradeoff between the complexity of the NN models and the accuracy is the most critical issue that confronts the NN designer. The higher the complexity is, the larger is the storage required to save the model and the heavier is the computational load. In fact, the high complexity is a natural result of the predominant ideology of building NN models based on sub-modules \cite{DeMattosNeto2017}.
The majority of researchers resort to using a number of simple NN modules to perform several operations on the available data, aiming to extract useful features and highlight the underlying relationships within the data, as done in \cite{Wang2020,Alhnaity2020,XuYue2019,Shi2019}. On the other hand, simple NN models cannot handle the multi-relationships, which could be present within the data.
In view of the characteristics of each problem, there is no general solution that fits all kinds of NN models. The majority of the solutions proposed in the literature are designed for a specific class of NN models or a particular problem. This research is concerned with multi-step ahead forecasting of periodic time-series information based on NNs.

It stands to reason that improving the accuracy without modifying the structure of the NN gives an additional advantage to the model by reducing the complexity as well as the computational load. That is the exact goal of this paper.
PID control is a well-known approach in the industrial arena. It has been applied for controlling the output of a system over time, which can be interpreted as the control of time-series events. PID controller is characterized by a small number of tunable parameters compared with the enormous number of parameters in machine learning methods. This feature can be exploited to improve the performance of the learning systems with a near-zero impact on the complexity of the system as well as a small extra computational load.

In this paper, we propose a novel technique inspired by the well-known PID control concept to control the output of NN-based systems, which are designed for multi-step ahead forecasting of periodic time-series information, particularly systems that depend on the iterative strategy to achieve multi-step ahead prediction. This method could be used as a supplementary method to support the NN model. The advantage of this method over the other methods is the negligible number of tunable parameters, where it adds only three tunable variables to the system.

This method is applied to two deep learning models proposed in \cite{Guo2018} and \cite{Salloom2021}. These models are used for multi-step prediction of water demand. The results are compared with the results of the same models reinforced with a supplementary processes to handle the accumulative error. Moreover, the PID-based method is applied to an LSTM-based model that used for power consumption prediction. The performance of the system including the PID-based method is compared with that of the same system without including the PID-based method.
The contribution of this paper can be summarized as follows:
\begin{itemize}
	\item A \ul{novel} method inspired by the PID control approach is designed that is able to improve the accuracy of a multi-step ahead time-series forecasting systems with insignificant impact on the system complexity and computational load.
	\item The designed method is applied to two water demand forecasting systems and another system for hourly forecasting of energy consumption. The efficiency is compared with another methods used to enhance the accuracy of the same systems.
\end{itemize}

The rest of this paper is organized as follows. Section \ref{Sec2} includes the methodology of designing and integrating the proposed method into the prediction system, while Section \ref{Sec3} includes the details of the case study of water demand prediction. The applicability of the proposed method to another time-series forecasting models is emphasized in section \ref{Sec4}. Comparison results and discussion are listed in Section \ref{Sec5}. The conclusion of this paper is stated in Section \ref{SecConc}.
\section{Methodology}
\label{Sec2}
This section includes the methodology of designing the PID-based method and the proper way to apply it to the system. Firstly, the designing strategy starts with the general formula of the PID control law that is used in the industrial field, each term of this formula is adapted individually according to the nature of the NN-based prediction system. The obtained PID-based method is able to enhance any NN-based prediction system that is meant for multi-steps ahead prediction of periodic data. Secondly, the right way to integrate the proposed method into the prediction system is discussed in detail. 
\subsection{The proposed method}\label{Sec2Sub1}
% \linenumbers
PID approach is used in industrial control systems to correct the system's output to follow a desired target or trajectory \cite{Rabah2019}. Assuming that the system is an NN model used to forecast $ n $ time-steps ahead of periodic data with a period of $ T $, the prediction results need to be controlled to follow the trajectory formed by the actual values of the data of $ n $ steps. Let $ PV(t) $ and $ RV(t) $ be the predicted value and the actual water demand value for time-step $ t $, respectively. The controller aims to make $ PV(t) $ follows the trajectory $ RV(t) $ where $ t = 1,2, ... , n $.
The NN-based prediction system is a discrete system.
The following equation represents the general equation of a PID controller for such a system \cite{Zhao2019a,Wang2019b}: 

\begin{equation}
\label{Eq1_PIDrule}
u(t) = - K_p e(t) -K_i \sum_{0}^{t} e(t) - K_d \big( \Delta e(t)\big)
\end{equation}
where $ e(t) $ represents the system error, $ K_p $ is a positive number that represents the proportional gain, $ K_i $ is a positive number expresses the integral gain, $ K_d $ is a positive number that represents the derivative gain. The final output $ P(t) $ of the system becomes as follows:

\begin{equation}
\label{Eq2_FinalOutput}
P(t) = PV(t) + u(t)
\end{equation}
The error of the system is as follows:

\begin{equation}
\label{Eq3_FinalError}
e(t) = P(t) - RV(t)
\end{equation}
Since the feedback is not available until the end of the prediction process of $ n $ steps, the error of the period $ t-T $ is used. The integral part changes to the summation of the prediction error of the time-steps before $ t-T $.  In order to keep the summation part limited, it is reset after the end of the prediction round, thus the error is summed up starting with the first step in the previous round, $ (i-1)T $, where $ i $ is the number of the current round of prediction. The derivative part at a time-step $ t $ becomes the difference between the error at step $ t-T $ and at step $ t-T-1 $. Equation (\ref{Eq1_PIDrule}) becomes as follows:

\begin{equation}
\label{Eq4_OurPIDrule} 
u(t) = -K_p e(t-T)-K_i \sum_{x=(i-1)T}^{x=t-T} e(x) - K_d \big(e(t-T) - e(t-T-1) \big)
\end{equation}
\subsection{The integration of the PID-based method into the prediction system}
\label{Sec2Sub2}
In this work, the prediction system is an NN model that is used for multi-step prediction. The NN model performs an $ n $ time-steps of prediction using the iterative strategy. It uses the predicted values at each time-step to predict the value of the next time-step.
The PID-based method is integrated into the system in a way to enhance the accuracy of the final predicted value. The system will be configured and used as follows:

Firstly, the NN model is trained on the available data for one-step prediction until it gives a reasonable accuracy. The PID-based method is not used during training.

\begin{figure}[!h]
	\centerline{\includegraphics[width = 9cm]{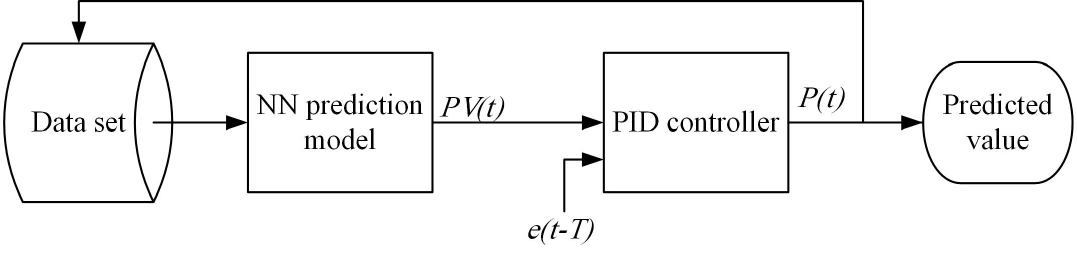}}%Not Found
	\caption{Prediction process.}
	\label{Fig0}
\end{figure}

Secondly, during prediction, the NN model achieves the prediction step by step; the output of the NN at every time-step is $ PV(t) $. The PID-based method designed in equations \eqref{Eq2_FinalOutput},\eqref{Eq3_FinalError}, and \eqref{Eq4_OurPIDrule} is applied to correct $ PV(t) $. The resulting value is the final forecasted value $ P(t) $ for time-step $ t $. $ P(t) $ is added to the database to be involved in the prediction of the next time-step $ t+1 $ if necessary. When real values become available, the current round's prediction errors are calculated to be used in the next round. Fig. \ref{Fig0} illustrates the system flow graph during prediction.

The error used in the first round of prediction should be initialized. There are three possible strategies to achieve that.
\begin{enumerate}
	\item[i.] Set the initial error to zero for all $ n $ steps. In this case, $ u = 0 $ and the controller becomes useless when applying it to the first $ n $ steps.
	\item[ii.] Set the initial error to the MAE that is obtained during the training, then the integral part increases rapidly due to the absence of negative errors. This increment may lead to instability.
	\item[iii.] The available data can be used to initialize the error; the NN model can be run to predict one round of $ n $ steps of the available data. In this round, the initial error for the first prediction step is set to zero, and the PID-based method is run over the resulted values. When running the system for the second round of prediction, the error from the first round can be used. Moreover, during the initialization process, the error of the previous prediction step $ t-1 $ in the first round of prediction can be involved instead of the error at $ t-T $ from the first round to guarantee faster convergence. Thus, equation \eqref{Eq4_OurPIDrule} takes the following form during the initialization process:
	
	\begin{equation}
	\label{Eq5_InitialPIDrule}
	u(t) = -K_p e(t-1)-K_i \sum_{x=0}^{x=t-1} e(x) - K_d \big(e(t-1) - e(t-2) \big)
	\end{equation}
	
	The third strategy is the most appropriate way and it is used in this work.
\end{enumerate}	

\section{Case study: water demand prediction}
\label{Sec3}
The limits of water resources on our planet become more and more apparent as the world's population increases and the daily use per population increases. It is, therefore, essential that water distribution should be managed very efficiently. This can be achieved effectively when a reasonable estimation of the water demand in the future is available.
The water demand forecasting methods can be classified based on the forecasting horizon \cite{Donkor2014}. The short-term horizon implies forecasting the water demand during short periods that range from 15 minutes to a few days in length \cite{Mala-Jetmarova2017}, while the medium-term horizon means forecasting the water demand for periods with a span of a month to one year. The long-term horizon implies forecasting water demand during a period longer than five years. Different factors influence the prediction for each horizon \cite{Balacco2017,Lu2018}. Factors such as pricing policy, the economic situation of the area, water conservation policies, population size, weather conditions, and the history of water demand influence the forecasting results of all horizons \cite{Romano2016}. In contrast, the history of water demand, the weather conditions, and the specific importance of the day have a considerable impact on the prediction results of the short-term horizon \cite{Romano2014}.

The case addressed in this work is the case of short-term prediction, specifically, daily prediction with a prediction period of 15 minutes.
Recently, the applicability of artificial NNs in this field has been investigated widely. The most appropriate structure of an NN model used for water demand forecasting relies on deep learning methods.
In this research, the methods GRUN and DCGRU proposed in \cite{Guo2018} and \cite{Salloom2021}, respectively, are used to prove the efficiency of the PID-based method. Where both works depend on the historical data for prediction and design their NN model based on the GRU, but the structure of each model is different. Guo et al. \cite{Guo2018} propose a supplementary  NN (SPNN) model to overcome the problem of error accumulation, while the DCGRU model includes a pre-processing step of data classification to mitigate this problem.
To prove the effectiveness of the proposed method, It is applied to the GRUN model and the overall performance is compared with the performance after applying the SPNN presented in the original work. Moreover, the PID-based method is applied to the DCGRU model and results are compared with the results obtained by applying the classification step.

Hereunder is a description of the prediction machinery including the data, the structure of both NNs, the SPNN model structure, and the application of the PID-based method to the NN prediction model.
\subsection{Water demand data}
\label{Sec3Sub1}
\begin{table}[h]
	\caption{Statistics of water demand data.}
	\begin{center}
		\begin{tabular}{lll}
			\hline
			\textbf{Statistics}  & \textbf{DMA1} &\textbf{DMA2}\\
			\hline
			Average demand ($\text{m}^3/15$ min) & $ 81.4 $ & $ 87.0 $\\
			Maximum demand ($\text{m}^3/15$ min) & $ 157.0 $	& $ 149.0 $\\
			Minimum demand ($\text{m}^3/15$ min) & $ 30.0 $ &	$ 40.0 $\\
			Standard deviation&	$ 24.4 $ &	$ 21.1 $\\
			Type of DMA  &	Residential	& Industrial\\
			\hline
		\end{tabular}
		\label{DataInfo}
	\end{center}	
\end{table}
The water demand data is real data collected from two distinct measurement areas (DMAs) in Changzhou in China. The population of the first area is around $ 13000 $, and a small number of commercial facilities. The second area, DMA2, has a population of $ 8500 $ in addition to $ 300 $ factories.

\begin{figure}[!h]
	\centerline{\includegraphics[width = 14cm]{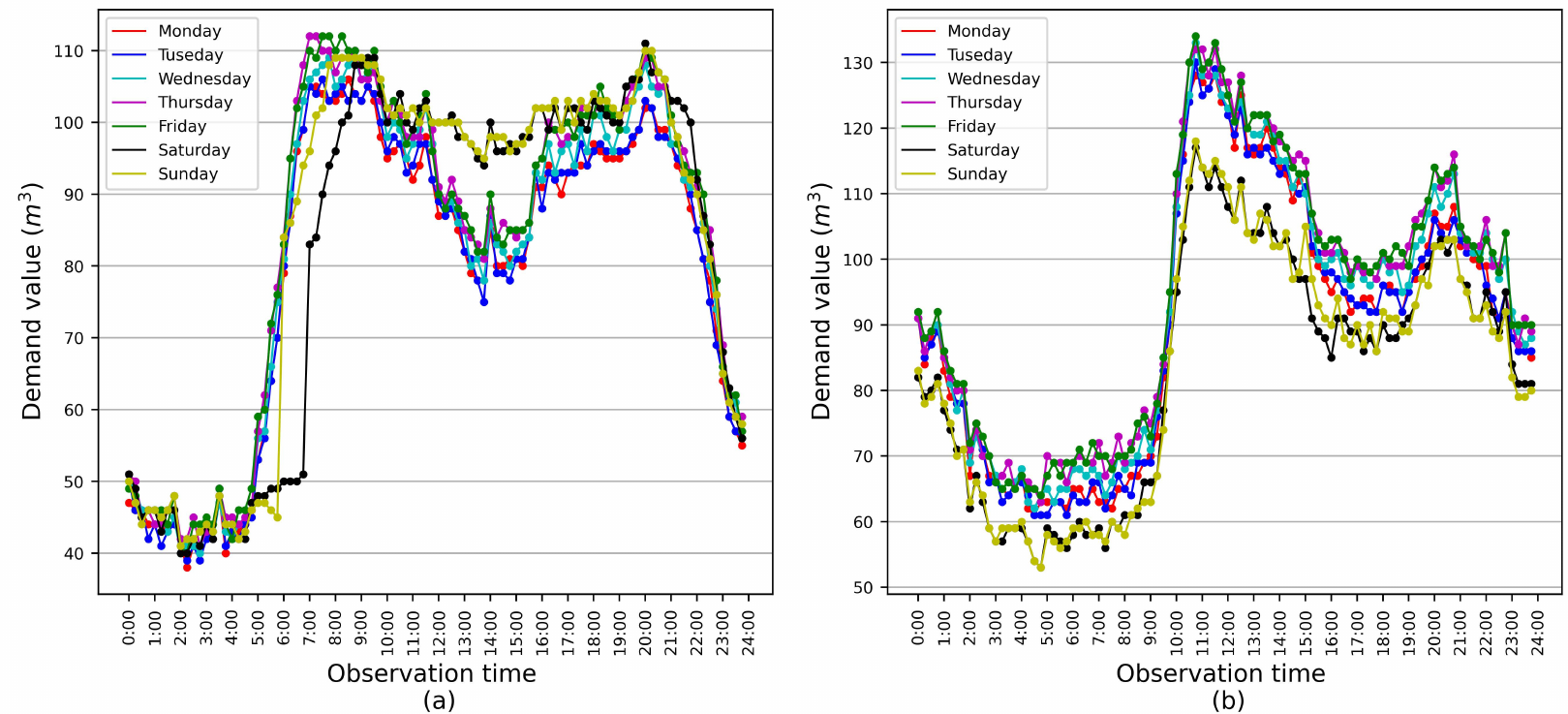}}
	\caption{Seven days of observations of water demand: (a) for DMA1; (b) for DMA2.}
	\label{Data}
\end{figure}

Fig. \ref{Data} illustrates one week samples of the two data sets. Fig. \ref{Data}(a) shows samples from DMA1 and Fig. \ref{Data}(b) shows samples from DMA2.  
Table \ref{DataInfo} demonstrates the statistics of the data. The data collected in $ 2016 $ for one year. Officials use an adaptive water management plan that requires information about the water demand every $ 15 $ minutes. Thus, the water demand is measured every $ 15 $ minutes, and the prediction period is set to the same length. The measured data transferred to the control room at the end of the day. This raises the need for a multi-step ahead prediction to do the water demand forecasting for $ 96 $ successive periods, $ 15 $ minutes each, to get the data required for a full day.

The dataset is divided into three sets, the training set, the validation set, and the testing set. Both the validation set and the testing set contain 10\% of the total data each, while the training set contains 80\% of the total data.
\subsection{The structure of the GRUN prediction model}
\label{Sec3Sub2}

\begin{table}[h]
	\caption{Selected water demand values for each feature.}
	\begin{center}
		\begin{tabular}{ll}
			\hline
			\textbf{Feature}& \textbf{Selected demand values} \\
			%\cline{2-4} 
			\hline
			Input sequence 1	& $V_{t-1}, V_{t-2}, V_{t-3}, V_{t-4}$, and $V_{t-5}$ \\
			Input sequence 2	& $V_{t-94}, V_{t-95}, V_{t-96}, V_{t-97}$, and $V_{t-98}$ \\
			Input sequence 3 & $V_{t-190}, V_{t-191}, V_{t-192}, V_{t-193}$, and $V_{t-194}$ \\
			\hline
		\end{tabular}
		\label{GRUNFeatureInfo}
	\end{center}
\end{table}
Guo et al. \cite{Guo2018} are of the opinion that historical data are sufficient for extracting the pattern of water demand. Disregarding the veracity of this claim, this work is based on the same GRU NN structure as it clearly shows the problem of error accumulation. The effectiveness of the PID-based method proposed in this work is shown in comparison to the supplementary model used by Guo et al. to enhance the prediction accuracy.

\begin{figure}[!h]
	\centerline{\includegraphics[width = 14cm]{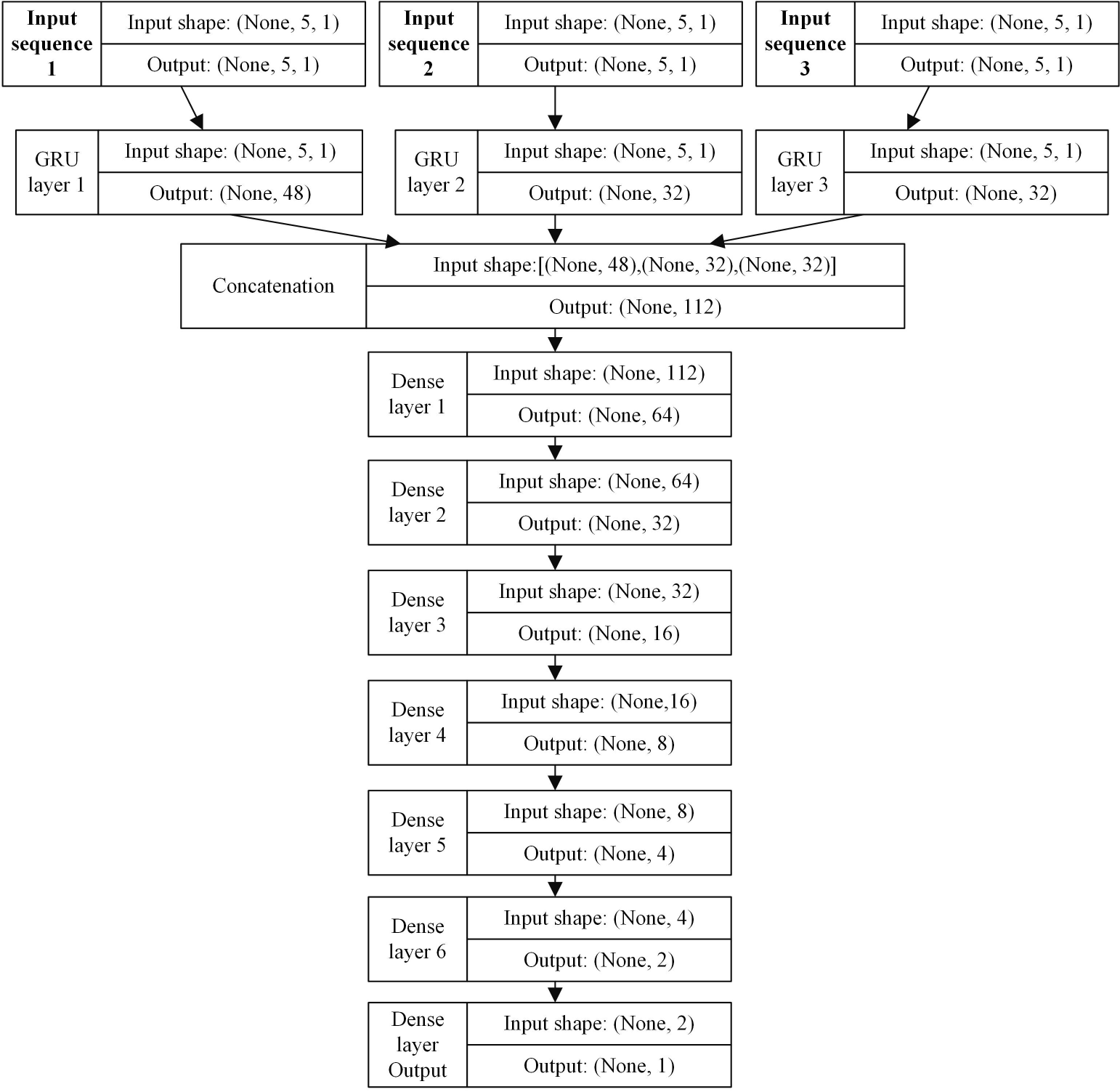}}%Not Found
	\caption{The structure of the GRUN model.}
	\label{fig_GRUN}
\end{figure}
The prediction model is built based on the GRU cell. Guo et al. assume that the prediction of the water demand at a period $ t $ depends on the close history and far history of water demand. They choose three data sequences as input for their model, five elements each. The first sequence represents the water demand in the most recent periods, which are the last five periods. The second one represents the water demand in the earlier periods. They choose five periods from the last day. While the third sequence represents the distant time, they are selected from the second day before the day of the period to predict. Table \ref{GRUNFeatureInfo} shows the elements of each sequence. 
The structure of the prediction model is shown in Fig. \ref{fig_GRUN}. It is composed of three GRU layers, and each layer extracts the sequential relationship between its input. The outputs of the three GRU layers are merged and sent to a block of six dense layers, which work upon the relationship between the three sequences. The output of the last dense layer is one value that represents the desired water demand during the period $ t $.
\subsection{The supplementary model (SPNN)}
\label{Sec3Sub3}
\begin{figure}[!h]
	\centerline{\includegraphics[width = 9cm]{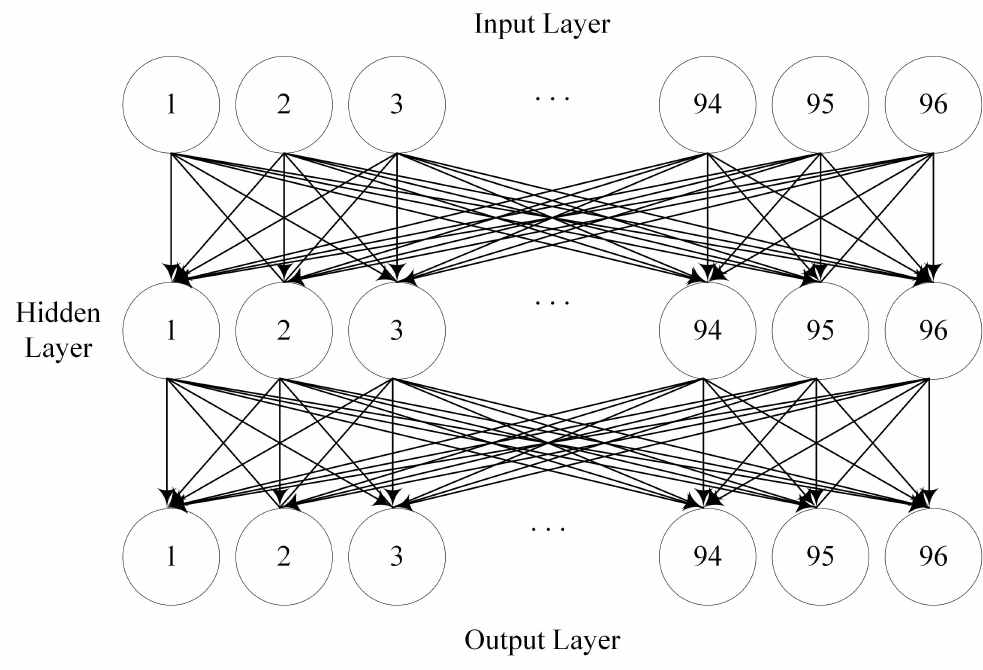}}%Not Found
	\caption{The structure of the SPNN model.}
	\label{fig_Correction}
\end{figure}

This model is proposed by Guo et al. \cite{Guo2018}. The goal is to bring the predicted water demand values closer to the real values. The SPNN model is a shallow NN that consists of an input layer and an output layer with one hidden layer. The input of this model is $ 96 $ predicted water demand values, while the output is also $ 96 $ water demand values as the correction of the predicted values. The structure of the model is demonstrated in Fig. \ref{fig_Correction}.

Regarding the application strategy of this model, firstly, the water demand for $ 96 $ periods should be forecasted using the prediction model, then the $ 96 $ values are sent as input to the SPNN model. The output values of the SPNN model are the final predicted values that are used in water plant management.
\subsection{The structure of the DCGRU model} \label{Sec3Sub4}
This model also depends on historical data to predict the demand for the next 15 minutes. The structure of the model includes a layer of 32 GRU units, with an output layer with one GRU unit. All data are classified into four classes, then 96 vectors, where each one contains a value of water demand with its belonging to each class, are sent to three fully connected layers to come up with 96 values, these values are the input to the GRU layer.

\begin{figure}[!h]
	\centerline{\includegraphics[width = 8cm]{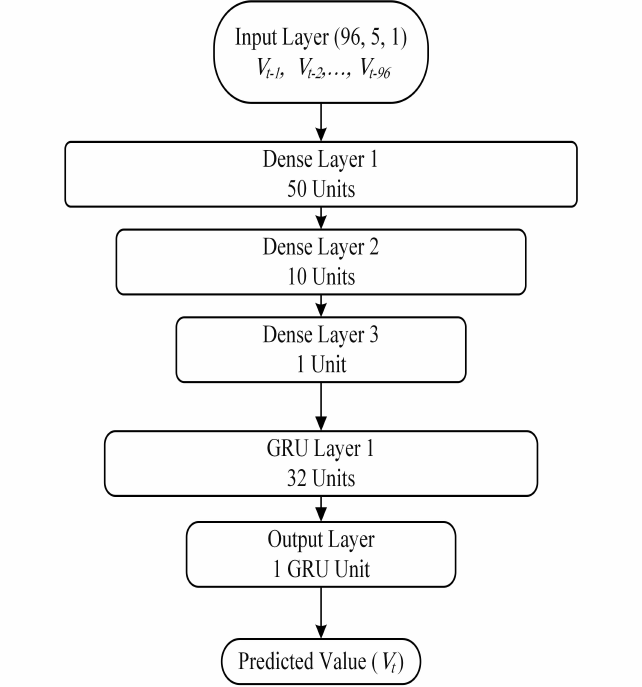}}%Not Found
	\caption{The structure of the DCGRU model.}
	\label{fig_DCGRU}
\end{figure}

The full structure of the model is shown in Fig. \ref{fig_DCGRU}. The main purpose of the classification step is to reduce the influence of the predicted value by creating a relationship between each value and other previous values. For more details about the DCGRU model, readers could refer to the original work in \cite{Salloom2021}. 

To evaluate our PID-based method, we apply the method to the DCGRU model instead of the classification step. Demand values $ t-1 $ to $ t-96 $ are sent to the GRU layer directly then apply the PID-based method to predict 96 steps. The final result is compared with the prediction result of DCGRU including the classification step.
\subsection{The application of the PID-based method for water demand prediction}
\label{Sec3Sub5}
For both GRUN and DCGRU models, the proposed PID-based method designed in \eqref{Eq4_OurPIDrule} is applied after forecasting each value in the series, and the resulted value is considered as the final prediction result. Then the value is added to the data to be used when forecasting the water demand for the next period. The error $ e(t) $, the integration of the error, and the derivative of the error are calculated based on that final prediction value. 

The critical problem in this method is to find the optimal values of the parameters of the controller, i.e. $ K_p, K_i, $ and $ K_d $. In fact, there are several methods to tune such a controller, such as deterministic methods as in \cite{Najafizadegan2017,Wang2017b}, evolutionary methods as in \cite{Salloom2020a}, or hybrid ones as in \cite{Feng2018}. In this work, the experimental way gives satisfactory results, where $ K_p $ is tuned firstly, then $ K_i $ and $ K_d $, in the beginning, $ K_p $ is set to a small positive number $ 0.1 $, then increased gradually with an incremental step of $ 0.1 $ until reaching the optimal value at which the smallest output error is obtained. Then $ K_i $ and $ K_d $ are set to small positive numbers starting with $ 0.0001 $ and an incremental step of $ 0.0001 $ as well. 
It has been found that the best control performance can be obtained when $ K_p, K_i, K_d \in [0, 1]$ and $ K_i, K_d < K_p $. The obtained values of $ K_p $,$  K_i $, and $K_d $ that used with the GRUN and DCGRU models are listed in Table \ref{K_values}. Although the experimental method of tuning the PID control law does not produce an optimal controller, it nevertheless results in an effective one.

\section{The applicability of the PID-based method to other time-series forecasting methods} \label{Sec4}
As we mentioned above, the applicability of the proposed method is guaranteed on time series data that satisfy the following: (i) the series should be periodic with a period of $ T $, to guarantee that the value at time point $ t $ close enough to that at time point $ t-iT $ where $ i = 1,2,3… $, (ii) The original prediction model is an NN model. (iii) Multi-step prediction strategy used is the iterative strategy. The applicability of the proposed method with other types of prediction models or with different multi-step prediction strategies will be studied in the future work.

In a distributed environment, the data will be distributed in many storages, the approach of federated learning will be used, where several copies of the NN model will be trained on several data sets, then any merging method could be used to merge the trained models, finally, the PID-based method is applied to the final model and any training data set could be used to tune the parameter of the PID law. Further investigation will be conducted in future work. 

In order to clarify the applicability of the proposed method, we apply it to enhance the prediction results of CNN-LSTM model in \cite{Yan2018p}, this model is used for multi-step prediction of hourly consumption of electricity.

\subsection{Electricity consumption data}
\label{Sec4Sub1}

\begin{figure}[!h]
	\centerline{\includegraphics[width = 15cm]{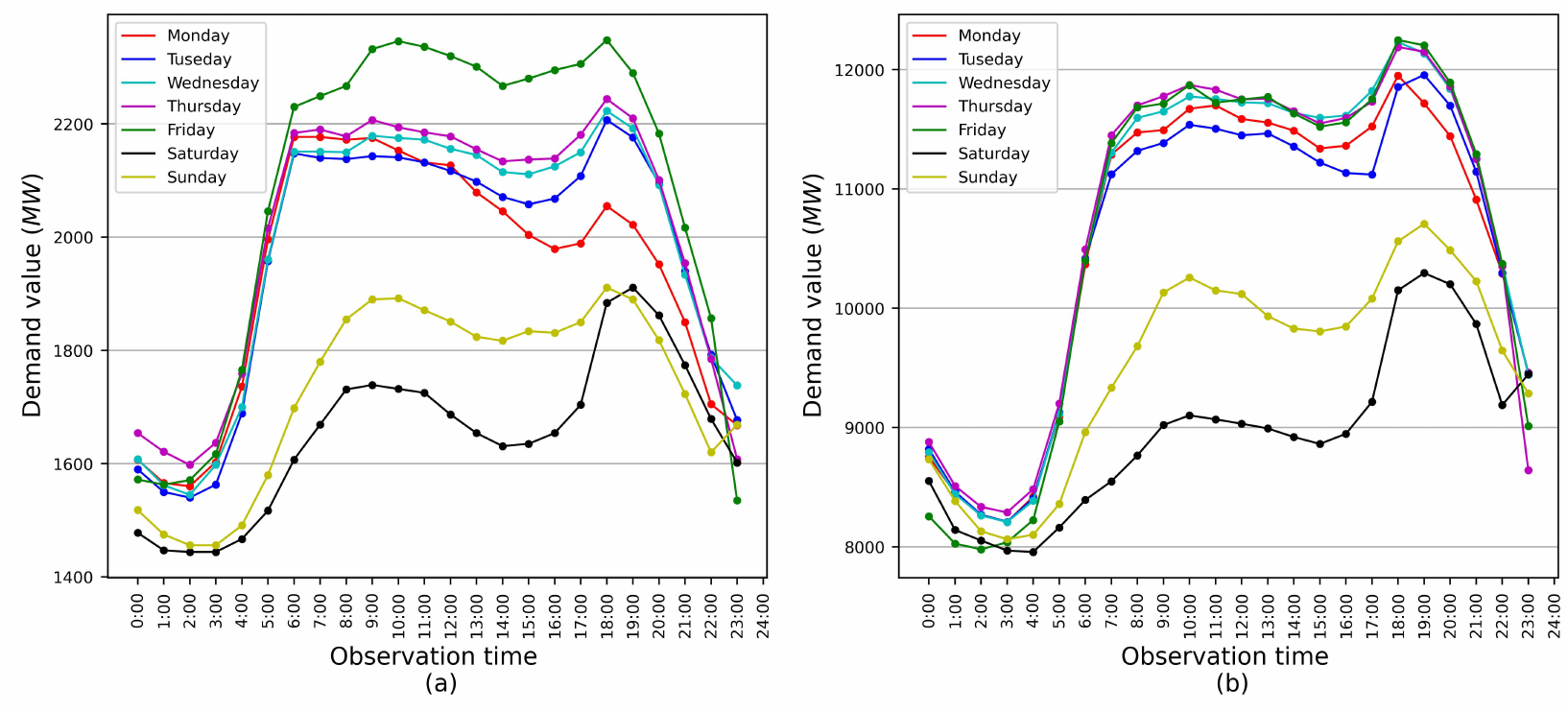}}
	\caption{Seven days of observations of electricity consumption in megawatt (mw/h): (a) for DAYTON data set; (b) for FirstEnergy data set.}
	\label{PowerData}
\end{figure}

\begin{table}[!h]
	\caption{Statistics of energy consumption data.}
	\begin{center}
		\begin{tabular}{lll}
			\hline
			\textbf{Statistics}  & \textbf{DAYTON} &\textbf{FirstEnergy}\\
			\hline
			Number of costumers & $ 500,000 $ & $ 6,000,000 $\\
			Number of reading & $ 121250 $ & $ 62850 $\\
			Average demand (mw/h)  & $ 2037.85 $ & $11701.68 $\\
			Maximum demand (mw/h)	& $ 3764 $	& $ 23631 $\\
			Minimum demand (mw/h) & $ 982 $ &	$ 7003 $\\
			Standard deviation &	$ 393.40 $ &	$ 2371.48$\\
			\hline
			mw = megawatt
		\end{tabular}
		\label{PowerStat}
	\end{center}	
\end{table}

The hourly consumption data set is taken from PJM Interconnection LLC, which is a regional transmission organization in the United States, it provides data from thirteen companies that serve different states in the USA. In this paper, only two data sets are chosen to test our method, first one is the DAYTON data set which contains energy demand supplied by the “Dayton Power and Light” company, which serve a customer base of 500 thousand people distributed in the area of West Central Ohio, including the area around Dayton, Ohio.
The second one is the FirstEnergy data set which contains power demand supplied by the “FirstEnergy” company, this company serve about 6 million clients in the area of Ohio, Pennsylvania, West Virginia, Virginia, Maryland, New Jersey and New York. DAYTON data sets contain observations from 1-10-2004 to 1-2-2018, while the FirstEnergy data set contains hourly observations from 1-6-2011 to 1-8-2018. Fig. \ref{PowerData} shows samples of one week of each data set, Fig. \ref{PowerData}(a) shows DAYTON data set, while Fig.\ref{PowerData}(b) shows FirstEnergy data set. Table \ref{PowerStat} lists some statistical information about each data set. These data sets are available for public use at  https://www.kaggle.com/robikscube/hourly-energy-consumption.

Both data sets are divided into three sets, $ 70\% $ used for training and $ 10\% $ for validation, while the rest is used for testing.

\subsection{The structure of the CNN-LSTM model}
\label{Sec4Sub2}
This model is proposed in \cite{Yan2018p} for multi-step prediction of hourly power consumption. The input of the model is a series of 24 consecutive observation values. These values are passed to two convolution layers for extracting more features of the data, the resulted values form a sequence that is handled by an LSTM layer. 
The output layer of the model is a fully connected layer with a linear activation function. 

\begin{figure}[!h]
	\centerline{\includegraphics[width = 8cm]{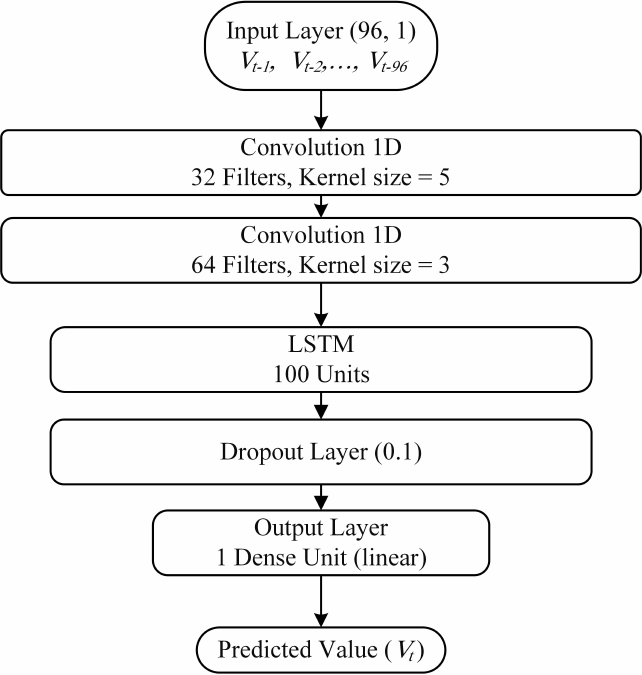}}%Not Found
	\caption{The structure of the CNN-LSTM model.}
	\label{CNN_LSTM_structure}
\end{figure} 

Fig. \ref{CNN_LSTM_structure} illustrates the structure of the CNN-LSTM model.
The activation function used in the convolutional layers is a ReLU, while for the LSTM layer, the authors use the default functions. .

\subsection{The application of the PID-based method to power prediction}
\label{Sec4Sub3}
CNN-LSTM model is trained on the training set of DAYTON and FirstEnergy data sets separately until reaching a prediction accuracy similar or higher than what has been obtained in the original work in \cite{Yan2018p}. As for water demand prediction, the PID booster is applied in the prediction phase. Firstly, power consumption is predicted by the CNN-LSTM model then the controller in equation (\ref{Eq4_OurPIDrule}) is applied, and its output is taken as the final output of the system. This final value is used in the input of the next prediction step. This process is repeated iteratively until predicting 24 steps. When real data is available, the prediction error is calculated to be used when predicting the next prediction round which contains 24 steps. the parameters of the controller are tuned using grid search. The value of $ K_p $,$  K_i $, and $K_d $ are listed in Table \ref{K_values}.

\begin{table}[h]
	\caption{The optimal values of the controller parameters used with each model.}
	\begin{center}
		\begin{tabular}{lllll}
			\hline
			\textbf{Data set}  & \textbf{NN Model} &\textbf{$ K_p $} &\textbf{$ K_i $} &\textbf{$ K_d $}\\
			\hline
			DMA 1 & GRUN & $ 0.4 $ & $0.01 $ & $0.001$\\
			DMA 2 & GRUN & $ 0.5 $ & $ 0.001 $ & $ 0.001$\\
			DMA 1 & DCGRU & $ 0.2 $ & $ 0.01 $ & $ 0.001$\\
			DMA 2 & DCGRU & $ 0.22 $ & $ 0.001 $ & $ 0.001$\\
			DAYTON & CNN-LSTM & $ 0.1$ & $ 0.0001 $ & $ 0.001$\\
			FirstEnergy & CNN-LSTM & $ 0.13 $ & $ 0.001 $ & $ 0.001$\\
			\hline
		\end{tabular}
		\label{K_values}
	\end{center}	
\end{table}
\section{Results and discussion}
\label{Sec5}
\subsection{Evaluation criteria}
\label{Sec5Sub1}
The effectiveness of the proposed PID-based method is investigated based on the prediction accuracy and the complexity of the forecasting system. Regarding water demand forecasting, two cases are compared, the first one is when the proposed PID-based method is applied to the GRUN model instead of the SPNN, and the DCGRU model instead of the classification method. While the second case is when the SPNN and the classification step are applied to the GRUN model and DCGRU model, respectively.

The prediction accuracy is measured by the mean absolute error (MAE) and the standard deviation (Std) of the error of the final result, in addition to the mean absolute percentage error (MAPE). Thus the distribution of the error can be estimated.

System complexity is measured based on (i) the computational load and (ii) the Akaike Information Criterion (AIC). \\
(i) Computational load is evaluated based on the execution time (ExT), including prediction time and data reshaping time. The training time is not measured since the method proposed in this work does not need training. ExT is measured for the process of predicting 96 steps, the measurement is repeated 1000 times; then the average is recorded. All process is executed using an ASUS laptop with core i7 processor and 16 GB of RAM and 1TB of hard disk in addition to 256 GB of SSD hard disk.
The model is built based on Keras library and Tensorflow backends with python 3.6 programming language.\\
(ii) AIC is a reliable tool for selecting the best between several competing statistical models that depend on the number of variables and output error of the models when applying the same dataset \cite{Paneiro2018,Seghouane2011}. The smaller the AIC, is the better. 
AIC for an NN model can be calculated based on the following equation \cite{Panchal2010a}:

\begin{equation} \label{Eq9}
AIC=n \ln{\bigg( \frac{RSS}{n} \bigg) } + 2w +\bigg( \frac{2w(w+1)}{n-w-1} \bigg)
\end{equation}
where $ w $ represents the number of the variables of the NN model, $ n $ represents the size of the dataset, $ RSS $ is the summation of the square of the forecasting error.
\subsection{Evaluation results of water demand prediction}
\label{Sec5Sub2}
\begin{table}[h]
	\caption{Evaluation results of water demand prediction.}
	\begin{center}
		\begin{tabular}{ p{2.8cm} p{0.7cm} p{1cm} p{1cm} p{1cm} p{1.2cm} p{1.5cm} p{1.4cm}}
			\hline
			\textbf{Feature} & \textbf{DMA} &\textbf{GRUN} &\textbf{GRUN + SPNN} &\textbf{GRUN + PID} &\textbf{DCGRU}& \textbf{DCGRU + classification}& \textbf{DCGRU + PID}\\
			\hline
			ExT(ms)&	$ 1 $ \newline $ 2 $ &	$ 78.516 $ \newline $ 78.516 $ &	$ 78.964 $ \newline $ 78.964 $&	$ 79.154 $ \newline $ 79.154 $ &$ 735 $ \newline $ 735 $ &$ 757 $ \newline $ 757 $&$ 773 $ \newline $ 773 $\\
			AIC  & $ 1 $ \newline $ 2 $ & $ 264606 $ \newline $ 263628 $ & $ 538116 $ \newline $ 527649 $ & $ 240945 $ \newline$ 242912 $&$ 51915 $ \newline $ 74956 $&$ 24325 $ \newline $ 30454 $& $ 25342 $ \newline $ 28344 $\\
			MAE($\text{m}^3/15$ min)& $ 1 $ \newline $ 2 $	& $ 3.82 $\newline $ 4.16 $ & $ 3.1 $ \newline $ 2.52 $& $ 2.76 $ \newline $ 2.8 $&$ 3.92 $ \newline $ 5.9 $ &$ 1.23 $ \newline $ 1.18 $&$ 1.92 $ \newline $ 2.19 $\\
			Std($\text{m}^3/15$ min) & $ 1 $ \newline $ 2 $ &	$ 3.32 $ \newline $ 3.56 $  & $ 2.78 $ \newline $ 2.34 $ &	$ 2.74 $ \newline $ 2.87 $&$ 2.72 $ \newline $ 2.9 $&$ 1.92 $ \newline $ 1.9 $&$ 1.8 $ \newline $ 1.91 $\\
			MAPE(\%) & $ 1 $ \newline $ 2 $ &	$ 5.14 $ \newline $ 4.98 $  & $ 3.79 $ \newline $ 2.83 $ &	$ 3.64 $ \newline $ 3.12 $&	$ 5.32 $ \newline $ 6.96 $&$ 1.52 $ \newline $ 1.43 $&$ 2.02 $ \newline $ 2.41 $\\
			\hline			
		\end{tabular}
	\end{center}
	\label{Evaluation}	
\end{table}
\begin{figure*}[!h]
	\centerline{\includegraphics[width = 11.5cm] {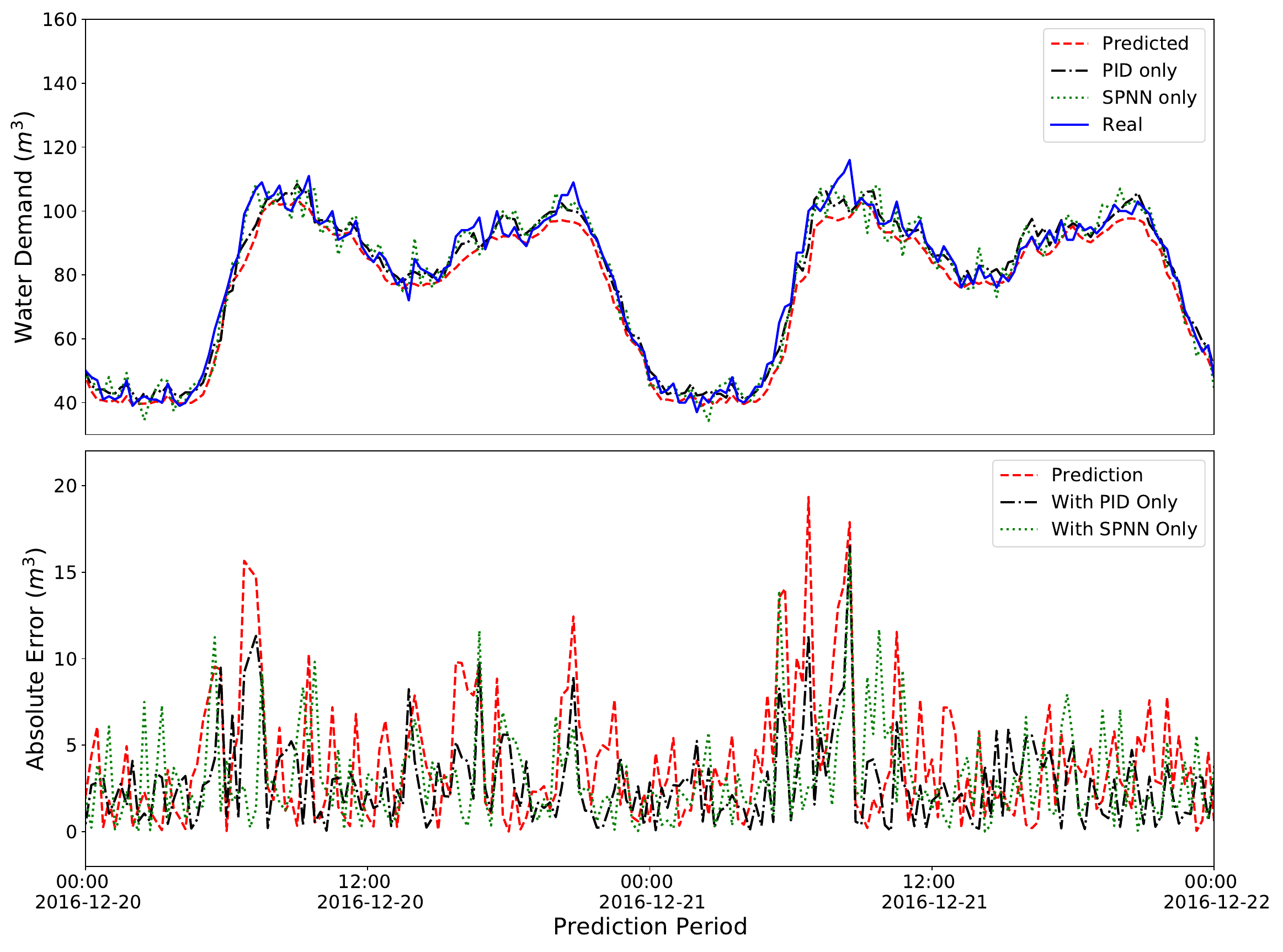}}%Not Found
	\caption{The output and the error of the prediction system when using different method to predict water demand in DMA1.}
	\label{fig_DMA1}
\end{figure*}
\begin{figure*}[!h]
	\centerline{\includegraphics[width = 11.5cm] {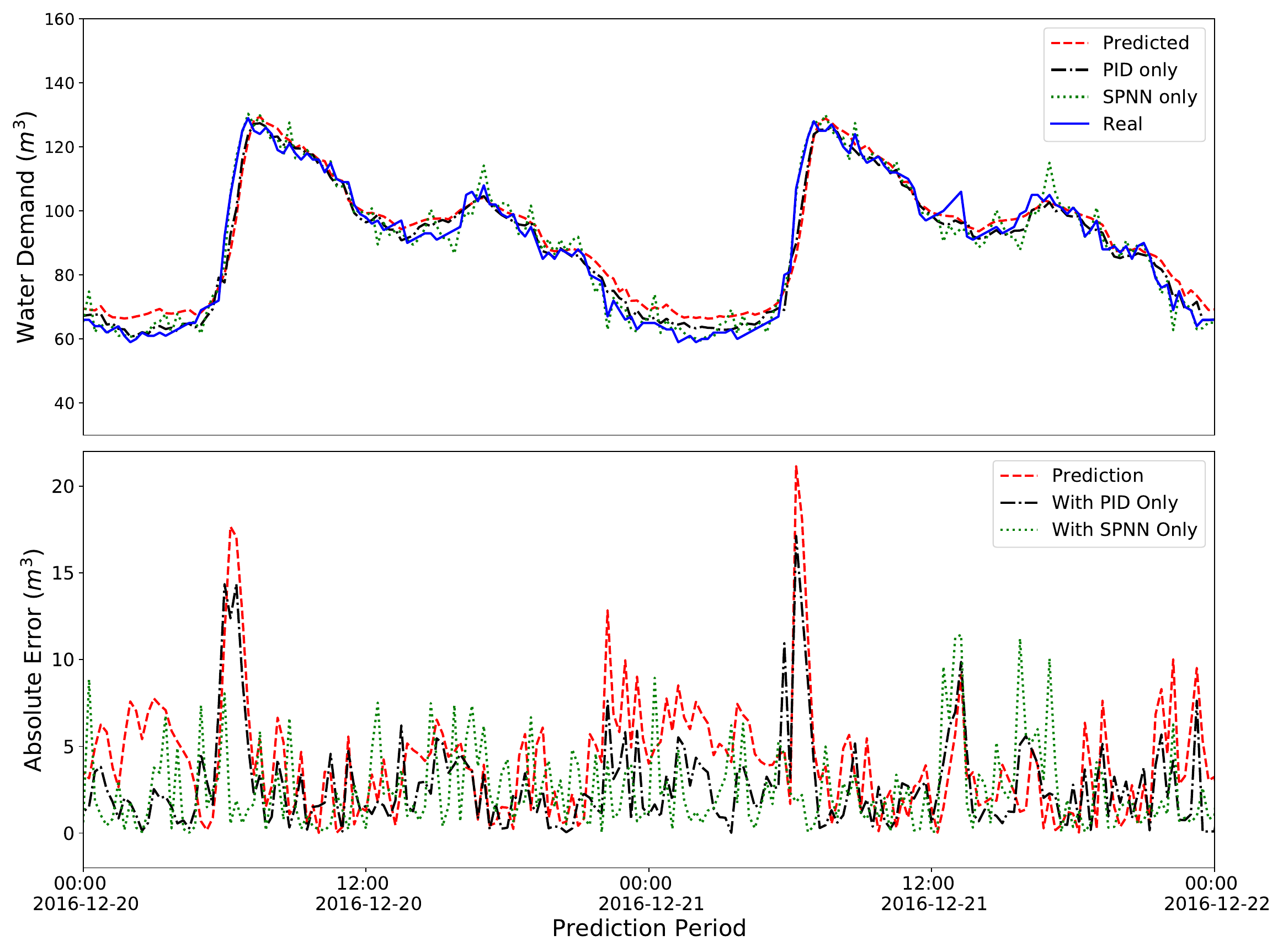}}
	\caption{The output and the error of the prediction system when using different method to predict water demand in DMA2.}
	\label{fig_DMA2}
\end{figure*}

The numerical evaluation results of both GRUN and DCGRU are stated in Table \ref{Evaluation}. 
\subsubsection{GRUN prediction system}
\label{Sec5Sub2Sub1}
The AIC of the original prediction NN model is $ 264606 $ and $ 263628 $ for DMA1 and DMA2, respectively. Applying the SPNN model doubles the AIC of the system to $ 538116 $ and $ 527649 $ for DMA1 and DMA2, respectively. On the contrary, applying the proposed method to the system reduces the AIC value of the system to $ 240945 $ and $ 242912 $ for DMA1 and DMA2, respectively.

Fig. \ref{fig_DMA1} and Fig. \ref{fig_DMA2} illustrate the prediction result and error for two days from the testing dataset for DMA1 and DMA2, respectively.
The MAE for the original prediction model is $ 3.82 \text{ m}^3/15$ min for DMA1 and $ 4.16 \text{ m}^3/15$ min for DMA2. Both the SPNN and the PID-based methods reduce the prediction error effectively. For DMA1, the PID-based method reduces the error to $ 2.76 \text{ m}^3/15 $ min, which is slightly less than $ 3.1 \text{ m}^3/15 $ min achieved by the SPNN model. On the contrary, for DMA2, the PID-based method reduces the MAE into $ 2.8 \text{ m}^3/15 $ min, which is higher than that is achieved by the SPNN model.

\begin{figure*}[!h]
	\centerline{\includegraphics[width = 12cm] {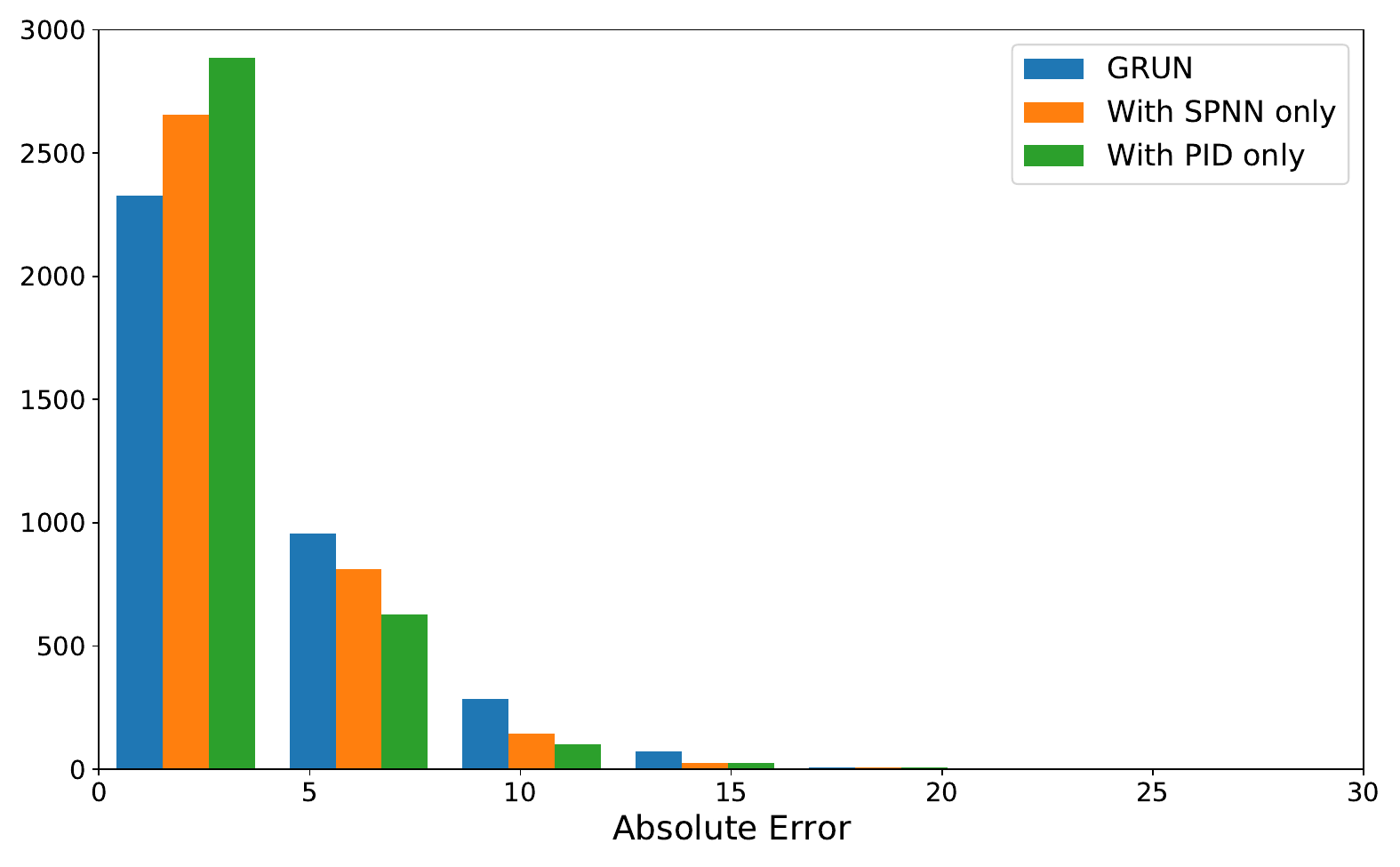}}%Not Found
	\caption{The histogram of the absolute error of the prediction system when using different method to predict water demand in DMA1.}
	\label{fig_Hist_DMA1}
\end{figure*}
\begin{figure*}[!h]
	\centerline{\includegraphics[width = 12cm] {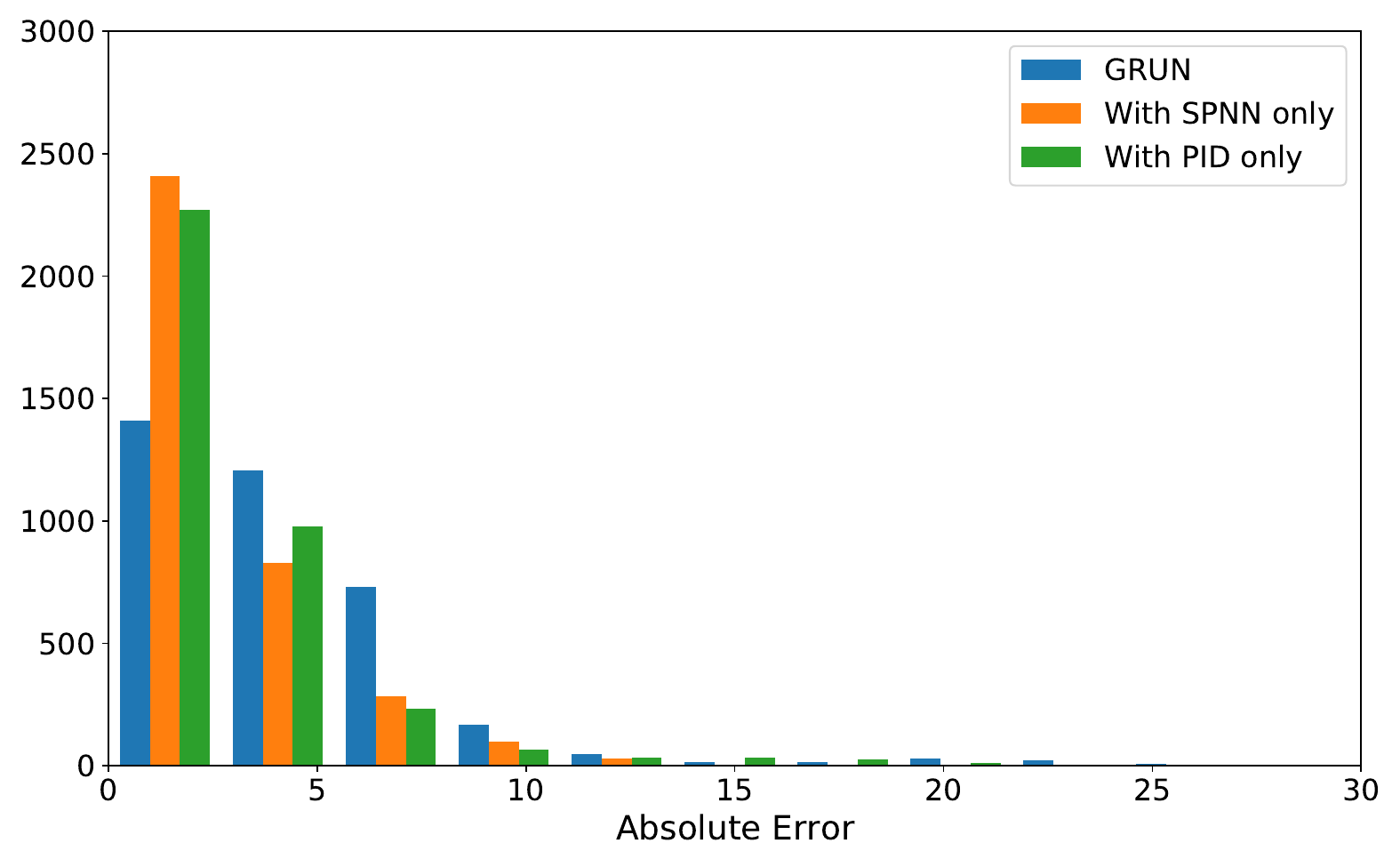}}
	\caption{The histogram of the absolute error of the prediction system when using different method to predict water demand in DMA2.}
	\label{fig_Hist_DMA2}
\end{figure*}

The histogram of the absolute error when using the SPNN model and the PID-based method to predict the testing data is shown in Fig. \ref{fig_Hist_DMA1} for DMA1 and Fig. \ref{fig_Hist_DMA2} for DMA2. It gives a better look at the error distribution. 

\subsubsection{DCGRU prediction system}
\label{Sec5Sub2Sub2}
\begin{figure*}[!h]
	\centerline{\includegraphics[width = 14cm] {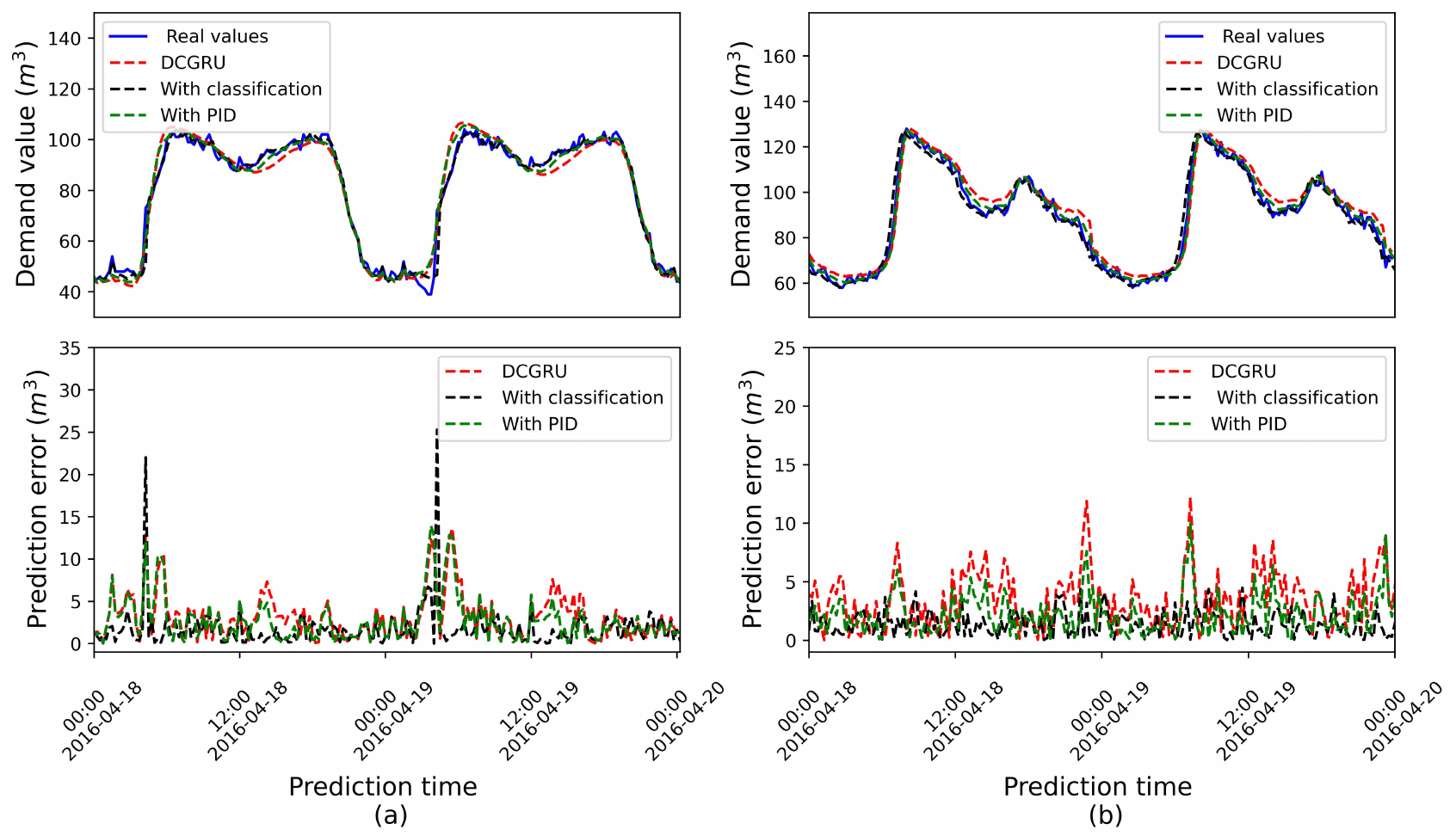}}
	\caption{The output and the error of the DCGRU prediction system: (a) for DMA1 data set; (b) for DMA2 data set.}
	\label{fig_PI_DCGRU}
\end{figure*}

The DCGRU model without classification nor PID represents the best model in term of tunable parameter, however, it achieves the highest MAE, which reaches $ 3.92 \text{ m}^3/15 $ min and $ 5.9 \text{ m}^3/15 $ min for DMA1 and DMA2, respectively. Thus, this model attains the highest AIC, $ 51915 $ for DMA1 and $ 74956 $ for DMA2. On the other hand, the classification step achieves the best accuracy with $ 1.18 \text{ m}^3/15 $ min and $ 1.38 \text{ m}^3/15 $ min of MAE for DMA1 and DMA2, respectively. However, applying the PID-based method to the DCGRU model reduces the MAE of the prediction system effectively, and reaches an AIC very close to what is obtained by the system with the classification step as shown in Table \ref{Evaluation}. Fig. \ref{fig_PI_DCGRU} shows the prediction results of two days of the water demand data sets when applying the PID-based method and the classification step, separately.
\subsection{Evaluation results of electricity consumption prediction}
\label{Sec5Sub3}
\begin{figure*}[!h]
	\centerline{\includegraphics[width = 14cm] {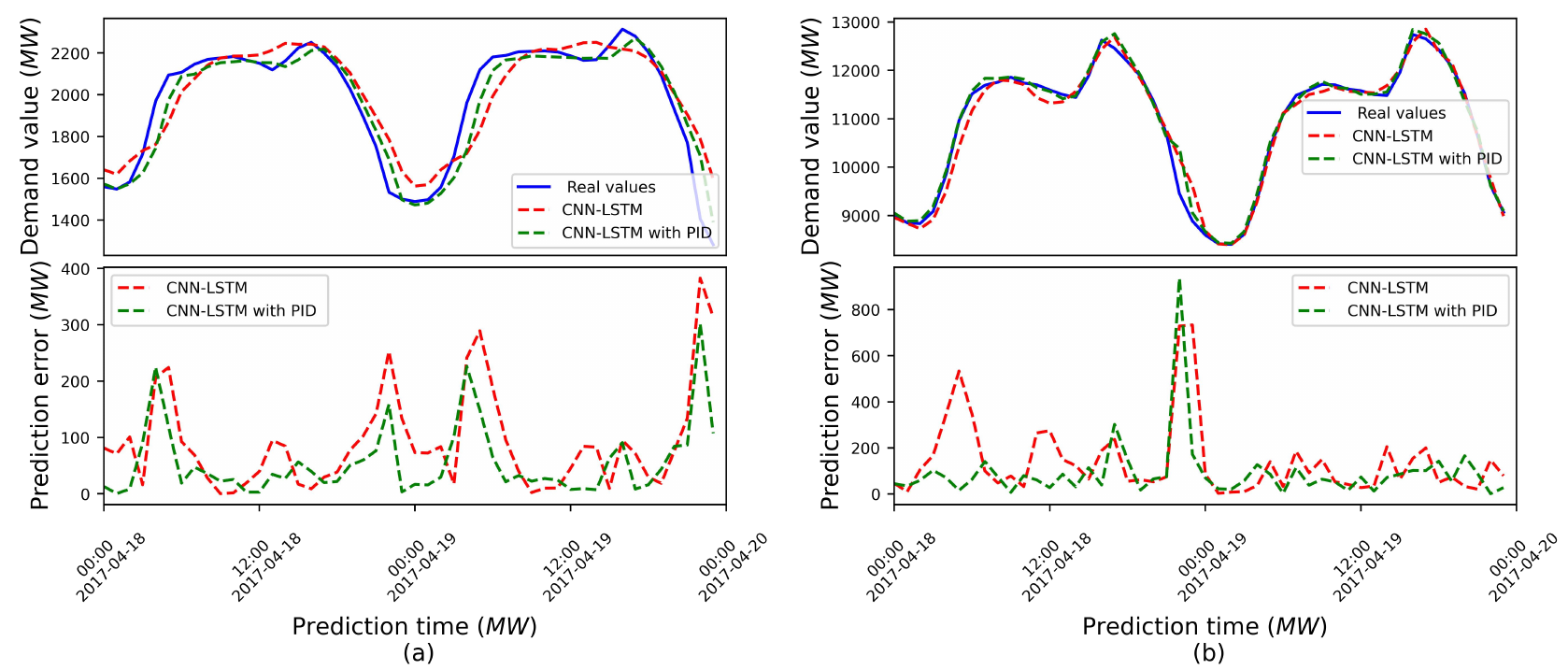}}
	\caption{The output and the error of the CNN-LSTM prediction system: (a) for DAYTON data set; (b) for FirstEnergy data set.}
	\label{fig_power}
\end{figure*}

Table \ref{EvaluationPower} describes all numerical results of the power prediction system. Fig. \ref{fig_power} shows the predicted values and the error for two days of data.
The AIC for the original system is $ 298070 $ and $ 154392 $ for DAYTON and FirstEnergy data sets, respectively. While after the application of the PID-based method, these numbers decline to $ 188848 $ and $ 143488 $. The proposed method contributes by reducing the MAE and MAPE of prediction. Without the PID-based method, the MAE values obtained are $ 109.27$ mw and $ 198.5$ mw for DAYTON and FirstEnergy data sets, respectively.

As for water demand prediction, ExT for power consumption prediction does not vary by data set. Logically, involving the PID-based method in the prediction process increases the execution time, where the ExT obtained before involving the PID is $ 170.21$ ms, while it rises to $ 315.83$ ms when using the PID-based method.

\begin{table}[h]
	\caption{Evaluation results of electricity consumption prediction.}
	\begin{center}
		\begin{tabular}{ p{2cm} p{2cm} p{2.2cm} p{2.5cm} }
			\hline
			\textbf{Feature} & \textbf{Data set} &\textbf{CNN-LSTM} &\textbf{CNN-LSTM with PID}\\
			\hline
			ExT(ms)& DAYTON  \newline FirstEnergy  &	$ 170.21$ \newline $ 170.21 $ &	$ 315.83 $ \newline $ 315.83 $ \\
			AIC  & DAYTON \newline FirstEnergy & $ 298070 $ \newline $154392  $ & $ 188848 $ \newline $143488 $\\
			MAE(mw)& DAYTON \newline FirstEnergy	& $ 109.27 $\newline $ 198.5 $ & $  71.07 $ \newline $132.4 $\\
			Std(mw) & DAYTON \newline FirstEnergy &	$ 143.15 $ \newline $ 319 $  & $ 94.63 $ \newline $231.03  $ \\
			MAPE(\%) & DAYTON \newline FirstEnergy & $ 5.6 $ \newline $ 1.74 $  & $ 3.47 $ \newline $ 1.16 $\\
			\hline			
		\end{tabular}
	\end{center}
	\label{EvaluationPower}	
\end{table}

\subsection{Discussion}
\label{Sec5Sub4}
This work targets the periodic time series data such as hourly or monthly water demand, power consumption or daily average temperature, etc. such kind of data are characterized by their seasonality, which means these data show a clear pattern that repeats itself every period of $ T $. Thus, the behaviour of the NN prediction model at time step $ t $ carries some information about the behaviour at time step $ t-T $.
Since the basic approach of the PID control implies driving the output of the system to a known target, however, in prediction problems, the target is unknown, therefore, the real error at time step $ t-T $ is used as a target instead of the unknown one at time $ t $.
For more understanding, substituting Equation \ref{Eq4_OurPIDrule} and Equation \ref{Eq2_FinalOutput} in Equation \ref{Eq3_FinalError} results in the following equation:

\begin{equation}
\label{Eq7_error}
e(t) = PV(t) - RV(t) -K_p e(t-T)-K_i \sum_{x=(i-1)T}^{x=t-T} e(x) - K_d \big(e(t-T) - e(t-T-1) \big)
\end{equation}

We have $ PV(t)-RV(t)= e_{NN }(t) $ is the prediction error of the NN model thus: 

\begin{equation}
\label{Eq8_error}
e(t) = e_{NN}(t) - K_p e(t-T) - K_i \sum_{x=(i-1)T}^{x=t-T}e(x) - K_d \big(e(t-T) - e(t-T-1) \big)
\end{equation}

The trained NN model guarantees a bounded prediction error such that $\Vert e_{NN}(t) \Vert \leq M$, where $ M \in \Re$ is a positive number.
The PID approach consists of three parts; the proportional part, $ K_pe(t) $, pulls the output of the system slightly towards the real value at $ t-T $. That is because, with periodic data, the neural network to have similar Behavior at steps $ t-iT $, where $ i=1,2,3… $, namely, the errors $ e(t-iT) $ are most likely to have the same sign. However, It is not guaranteed that $ RV(t)=RV(t-T) $ precisely, therefore, it is not wise to choose $  K_p=1 $.

In the industrial field, the PID controller is applied recursively until the output reaches the desired target, however, in the prediction problem, the PID-based method is applied once due to the uncertainty of the target, thus it is preferred to choose $ K_p < 1 $.

\begin{figure}[!h]%
	%\centering%
	\subfigure[][]{%
		\label{fig_Int-a}%
		\includegraphics[width = 7cm ,height=5cm] {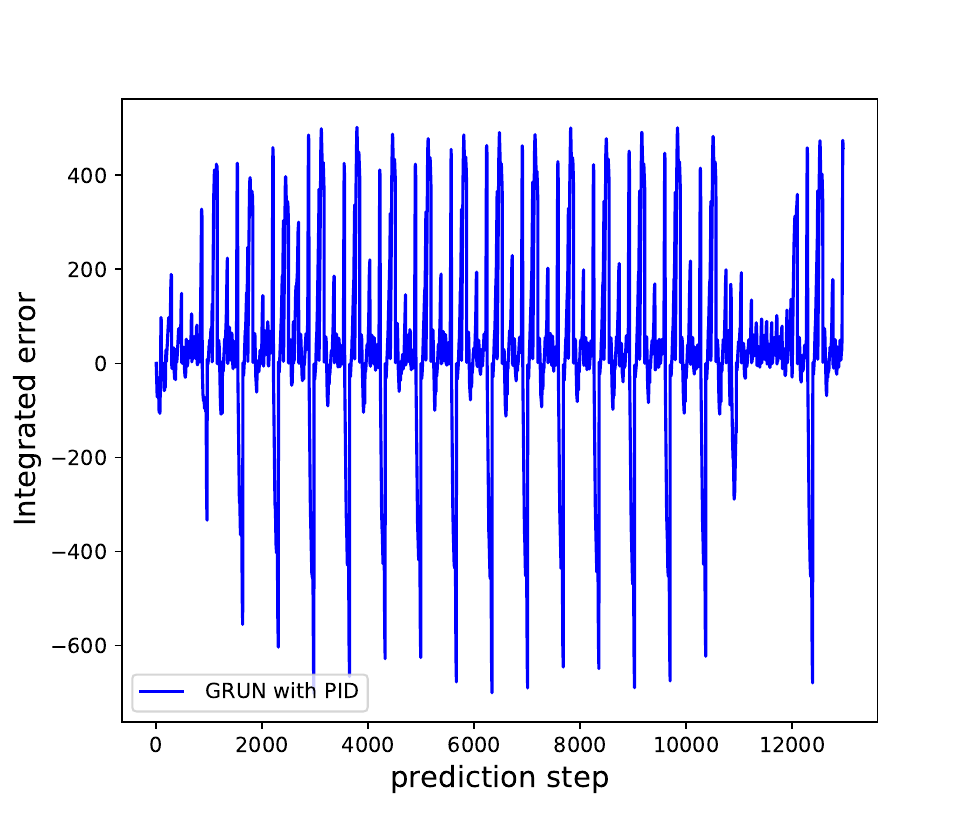}}%
	\hspace{4pt}%
	\subfigure[][]{%
		\label{fig_Int-b}%
		\includegraphics[width = 7cm,height=5cm] {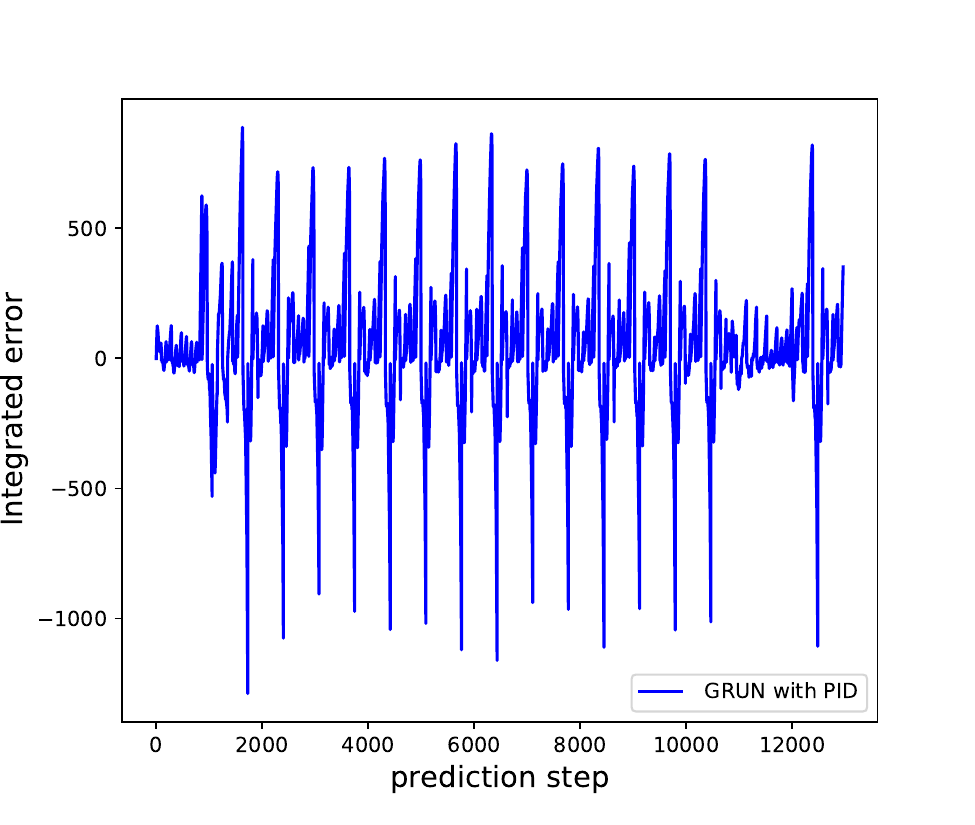}}\\
	\subfigure[][]{%
		\label{fig_Int-c}%
		\includegraphics[width = 7cm,height=5cm] {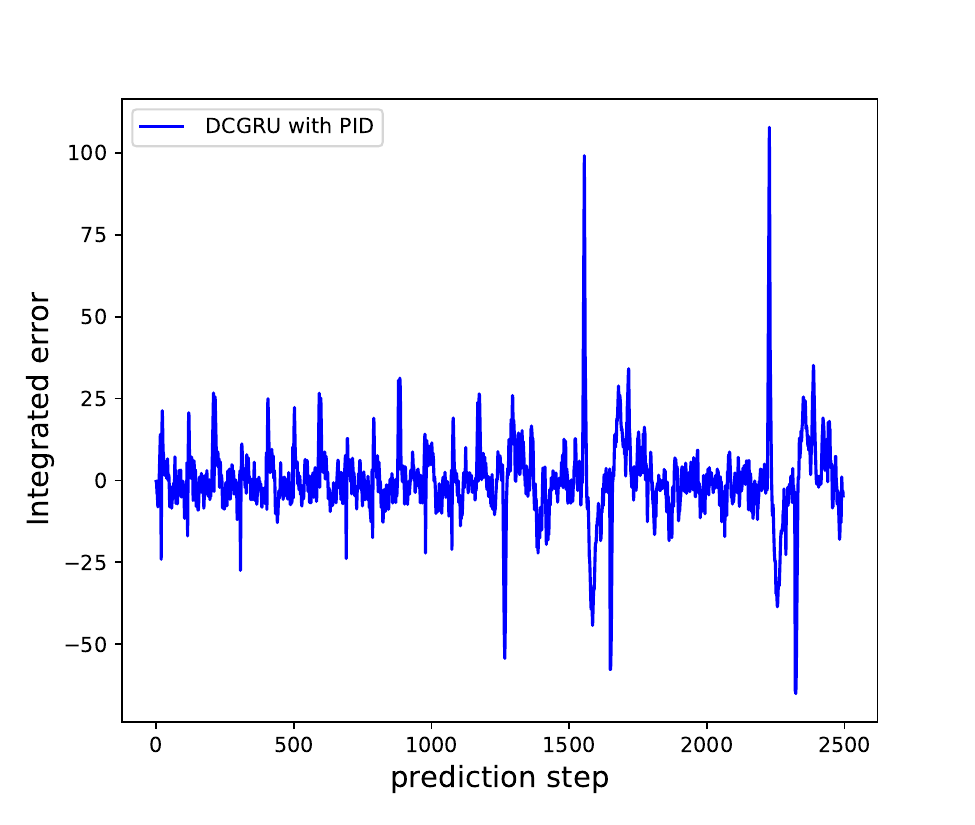}}%
	\hspace{4pt}%
	\subfigure[][]{%
		\label{fig_Int-d}%
		\includegraphics[width=7cm,height=5cm] {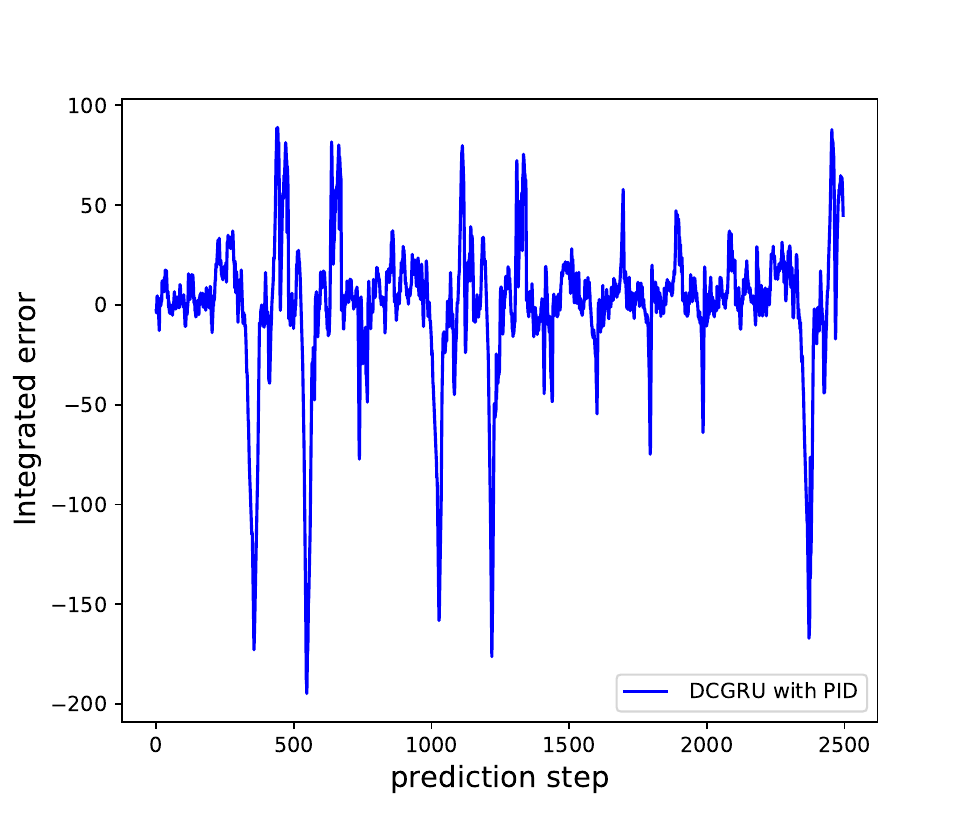}}\\
	\subfigure[][]{%
		\label{fig_Int-e}%
		\includegraphics[width = 7cm,height=5cm] {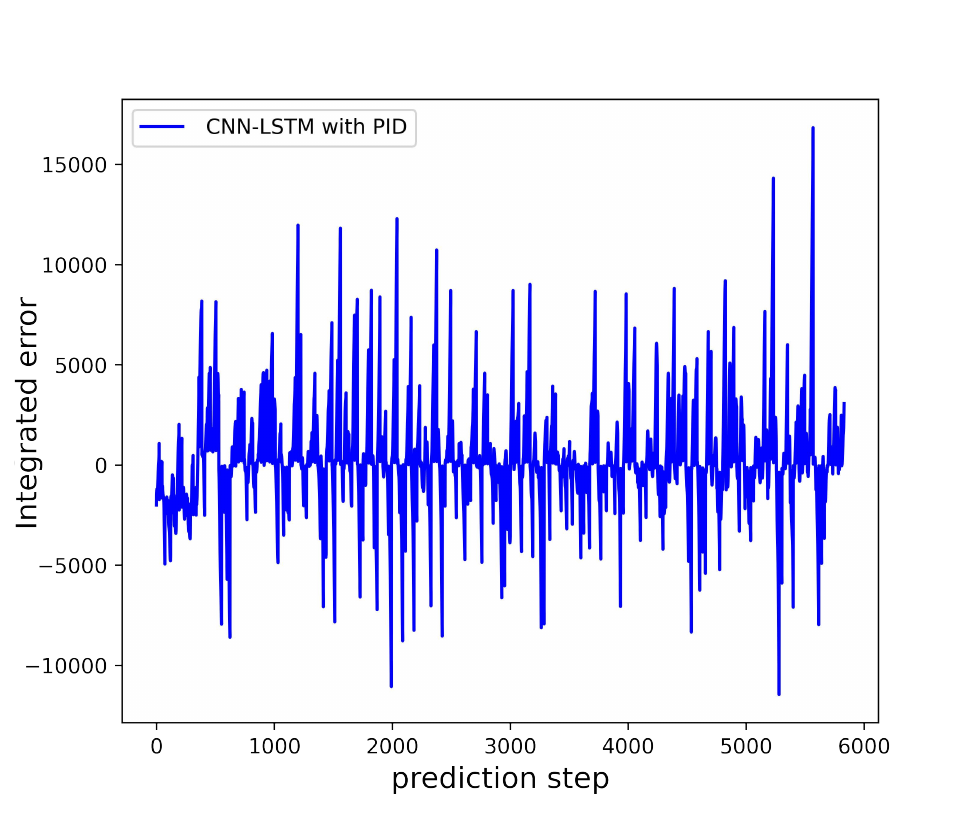}}%
	\hspace{4pt}%
	\subfigure[][]{%
		\label{fig_Int-f}%
		\includegraphics[width = 7cm, height=5cm] {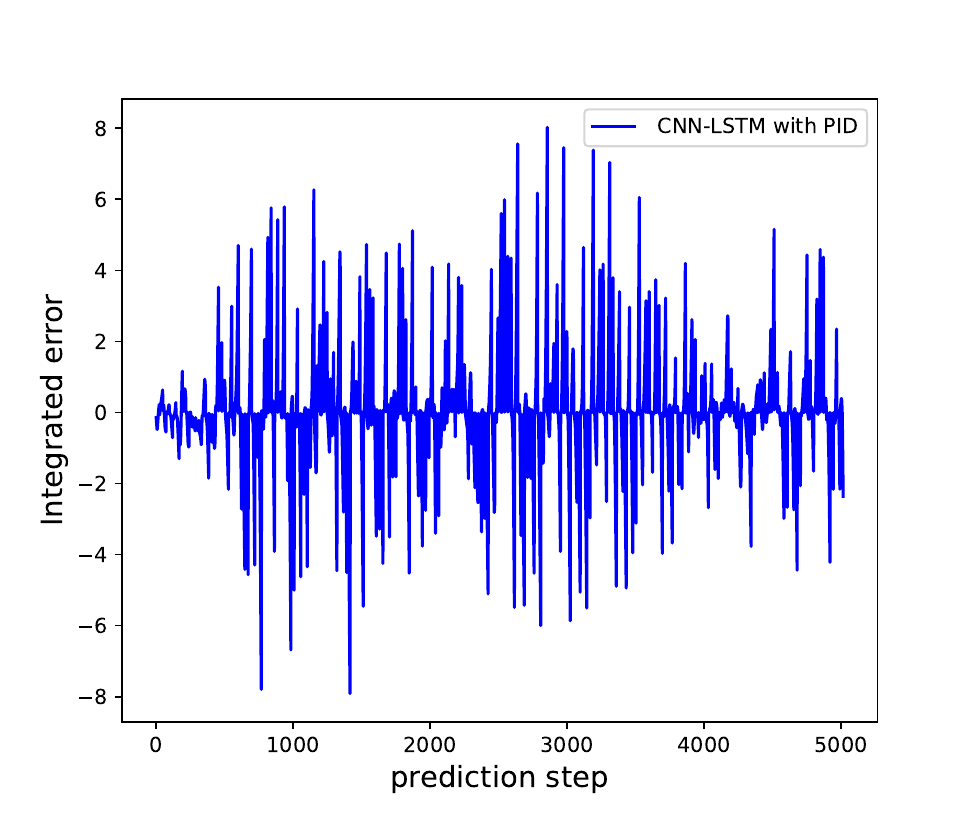}}%
	\caption[A set of four subfigures.]{The change of the summation term in equation \ref{Eq4_OurPIDrule} when applying the PID-based method to several models with different data sets:%
		\subref{fig_Int-a} GRUN model with DMA1;%
		\subref{fig_Int-b} GRUN model with DMA2;%
		\subref{fig_Int-c} DCGRU model with DMA1;%
		\subref{fig_Int-d} DCGRU model with DMA2;%
		\subref{fig_Int-e} CNN-LSTM model with DAYTON; and,%
		\subref{fig_Int-f} CNN-LSTM model with FirstEnergy.}%
	\label{fig_Int}%
\end{figure}

The summation part provides some information about the performance of the NN model, where the positive sum implies that the NN prediction error has been positive for several consecutive steps, $ PV>RV $, and vice versa. When the errors of several consecutive prediction steps have the same sign, the accumulative error increases.
The sum part aims to give the output of the NN model a big push to bring the output of the system to the other side of the real value $  RV(t) $, to guarantee that the input of the next prediction steps is distributed around the real values, thus the accumulative error is mitigated.
The prediction error follows the normal distribution, which means that the error takes both negative and positive values with similar probability, this, in turn, makes the sum of the errors oscillate around one point as shown in Fig. \ref{fig_Int}

The derivative part works as a brake to slow down the sum part while the consecutive error values still have the same sign but the altitude is getting smaller.

It is evident from Table \ref{Evaluation} that the use of the PID-based method brings reasonable improvements compared to other methods used in the literature to improve the accuracy of the NN-based prediction model. However, it results in the least complexity. The designed PID-based method adds only three variables to the prediction system: $ K_p $, $ K_i $, and $ K_d $. On the other hand, it reduces the error in a way that makes the AIC smaller than that of the other methods. Although the SPNN model reduces the error sufficiently; however, it increases the number of variables by $ 9312 $, which in turn leads to a large AIC. The classification step used with the DCGRU model results in a value of AIC similar to that obtained when applying the PID-based method to the same system. However, when there is some tolerance in the prediction accuracy, the PID-based method is preferred, while it is better to apply the classification step in other cases, especially when considering the prediction time, where the PID-based method has a longer prediction time than that of the other methods.

The histogram of the absolute error in Fig. \ref{fig_Hist_DMA1} and Fig. \ref{fig_Hist_DMA2} clarify that most of the error values fall in the intervals around or less than $ 5 \text{ m}^3/15$ min when using the SPNN and the PID-based methods, which confirm the usefulness of the two methods. However, the PID-based method reduces the number of error values that fall in an interval higher than $ 5 \text{ m}^3/15$ min more than what the SPNN model does.

ExT of the system with the PID-based method is slightly higher than that of the system with the SPNN model or with the classification method. This is due to the extra time required to calculate the integration of errors in the PID-based method. The ExT is identical for all data sets because the number of calculations needed to complete the process does not depend on the size of the data. 
\section{Conclusion}
\label{SecConc}
This work has proposed an effective method to boost the efficiency of NNs models that are designed for multi-step forecasting of time-series events based on the iterative prediction strategy. The proposed method depends on the concept of PID control, which is used widely for controlling time-series events in industrial fields. Tow NN models for water demand prediction and one for energy consumption prediction have been used to justify the efficiency of the PID-based method. Real water demand data from two areas in China and another two energy consumption data sets from the USA have been used. The prediction system when using the proposed PID-based method, have been compared with the same system when involving other methods to enhance the prediction error. The comparison results can be summarized as follows:
\begin{itemize}
	\item The PID-based method reduces the prediction error to the same level as that obtained by other methods considered in this study.
	\item The PID-based method increases the execution time slightly compared to the other methods.
	\item In terms of system complexity, the PID-based method demonstrates superior performance since it reduces the error effectively with a negligible effect on the number of variables in the system. On the contrary, the SPNN and classification methods substantially increase the number of variables in the system.
\end{itemize}
In the future, the applicability of the proposed method will be investigated when data are not periodic, as well as when using the multi-input multi-output strategy for multi-step prediction. Moreover, the optimization problem of the PID parameter will be studied in future works.
\bibliography{PIDNN2020}
% biographies
%\newpage
\vfill

\begin{wrapfigure}{l}{0.17\textwidth}
	\centering
	\vspace{-10pt}
	\includegraphics[width=0.17\textwidth]{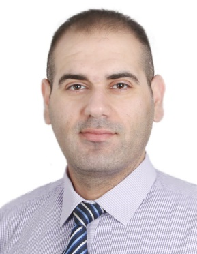}
	\vspace{-10pt}
	\hspace{-15pt}
\end{wrapfigure}
\noindent\textbf{Tony Salloom}
received the B.Sc. in electronics and computer engineering from Aleppo University, Aleppo, Syria, in 2008, and the M.Eng. in information and telecommunication engineering from the University of Science and Technology Beijing, Beijing, China, in 2016. \\
From 2009 to 2013, he served as a Database Engineer at the Syrian Telecommunication Company, Aleppo, Syria. He is currently pursuing the Ph.D. degree with the School of Automation and Electrical Engineering, University of Science and Technology Beijing, Beijing, China. His current research interests include Deep learning, time-series forecasting, intelligent control systems, and robotics.
\newline

\begin{wrapfigure}{l}{0.17\textwidth}
	\centering
	\vspace{-10pt}
	\includegraphics[width=0.17\textwidth]{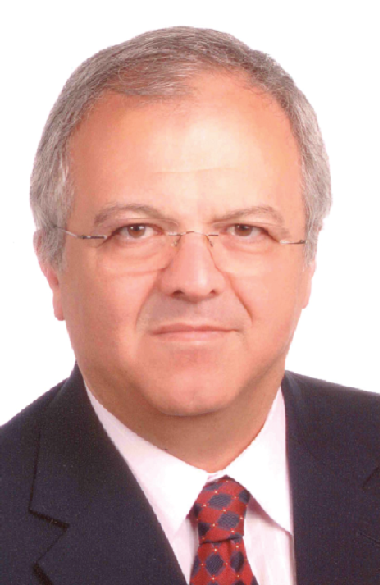}
	\vspace{-10pt}
	\hspace{-15pt}
\end{wrapfigure}
\noindent\textbf{Okyay Kaynak}
received the B.Sc. degree with first class honors and Ph.D. degrees in electronic and electrical engineering from the University of Birmingham, UK, in 1969 and 1972 respectively. From 1972 to 1979, he held various positions within the industry. In 1979, he joined the Department of Electrical and Electronics Engineering, Bogazici University, Istanbul, Turkey, where he is currently a Professor Emeritus, holding the UNESCO Chair on Mechatronics. He is also a 1000 People Plan Professor at University of Science \& Technology Beijing, China. He has hold long-term (near to or more than a year) Visiting Professor/Scholar positions at various institutions in Japan, Germany, U.S., Singapore and China. His current research interests are in the fields of intelligent control and industrial AI applications. He has authored four books, edited five and authored or co-authored more than 400 papers that have appeared in various journals and conference proceedings. Dr. Kaynak has served as the Editor in Chief of \emph{IEEE Trans. on Industrial Informatics and IEEE/ASME Trans. on Mechatronics} as well as Co- Editor in Chief of \emph{IEEE Trans. on Industrial Electronics}. Additionally, he is on the Editorial or Advisory Boards of several scholarly journals. In 2016, he received the Chinese Governments Friendship Award and Humboldt Research Prize. Most recently, in 2020, he received the International Research Prize of the Turkish Academy of Sciences.

Dr. Kaynak is active in international organizations, has served on many committees of IEEE and was the president of IEEE Industrial Electronics Society during 2002-2003. He was elevated to IEEE fellow status in 2003. 
\newline

\begin{wrapfigure}{l}{0.17\textwidth}
	\vspace{-14pt}
	\includegraphics[width=0.17\textwidth]{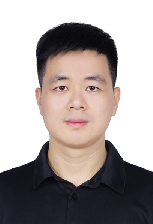}
	\vspace{-19pt}
\end{wrapfigure}
\noindent\textbf{Xinbo Yu He}
(S'16-M'20) received the B.E. degree in control technology and instrument from the School of Automation and Electrical Engineering, University of Science and Technology Beijing, Beijing, China, in 2013 and the Ph.D. degree in control science and engineering from the School of Automation and Electrical Engineering, University of Science and Technology Beijing, Beijing, China, in 2020. He is currently working as an associate professor in the Institute of Artificial Intelligence, University of Science and Technology Beijing, Beijing, China. His current research interests include adaptive neural networks control, robotics and human-robot interaction.
\newline

\begin{wrapfigure}{l}{0.17\textwidth}
	\vspace{-14pt}
	\includegraphics[width=0.17\textwidth]{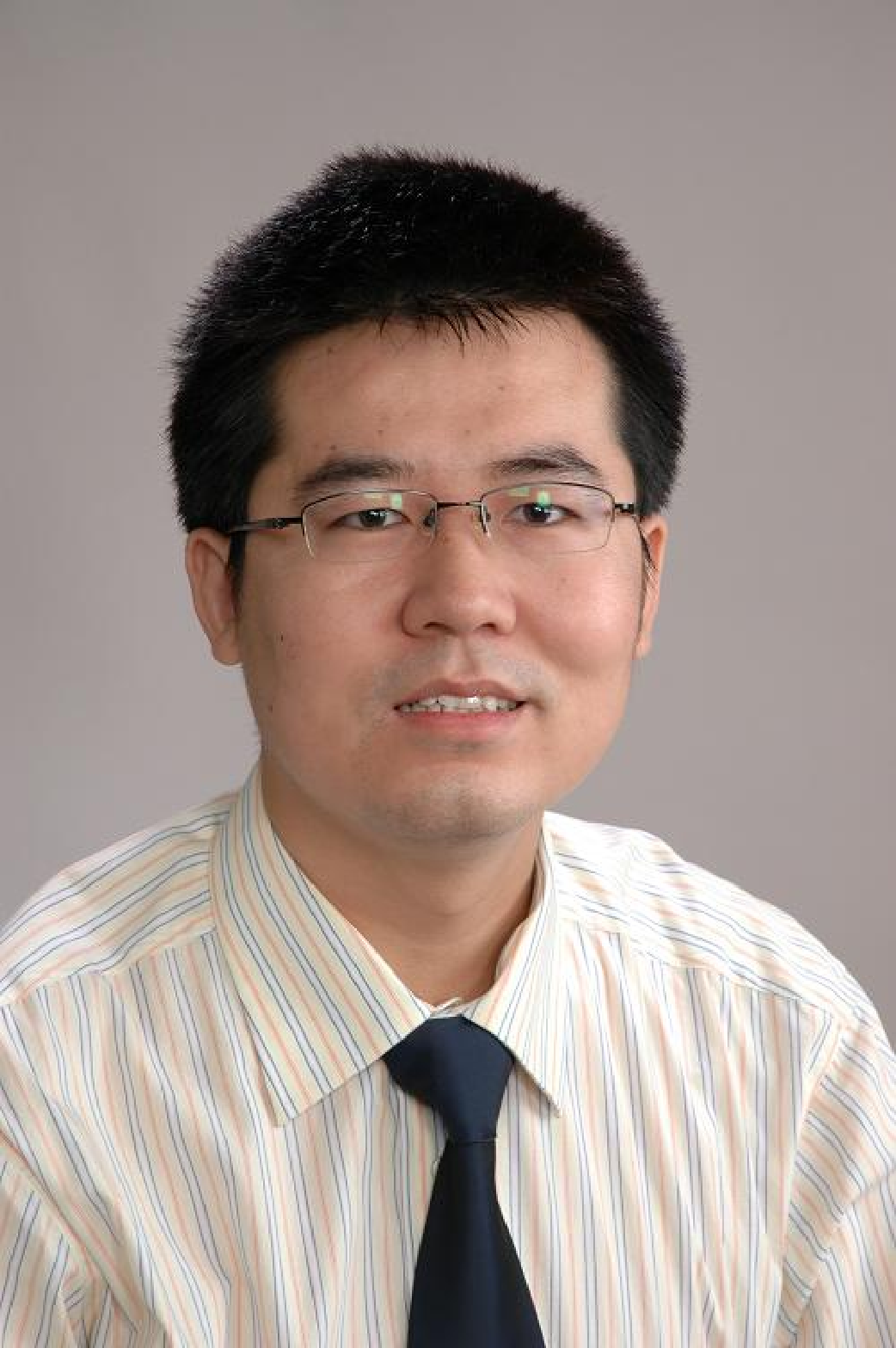}
	\vspace{-19pt}
\end{wrapfigure}
\noindent\textbf{Wei He}
(S'09-M'12-SM'16) received his B.Eng. in automation and his M.Eng. degrees in control
science and engineering from College of Automation Science and Engineering, South China University of Technology (SCUT), China, in 2006 and 2008, respectively, and his Ph.D. degree in control
science and engineering from from Department of Electrical \& Computer Engineering, the National University of Singapore (NUS), Singapore, in 2011.

He is currently working as a full professor in School of Automation and Electrical Engineering, University of Science and Technology Beijing, Beijing, China. He has co-authored 2 books published in Springer and published over 100 international journal and conference papers. He was awarded a Newton Advanced Fellowship from the Royal Society, UK in 2017. He was a recipient of the IEEE SMC Society Andrew P. Sage Best Transactions Paper Award in 2017. He is serving the Chair of IEEE SMC Society Beijing Capital Region Chapter. He is serving as an Associate Editor of \emph{IEEE Transactions on Robotics}, \emph{IEEE Transactions on Neural Networks and Learning Systems}, \emph{IEEE Transactions on Control Systems Technology}, \emph{IEEE Transactions on Systems, Man, and Cybernetics: Systems}, \emph{IEEE/CAA Journal of Automatica Sinica}, \emph{Neurocomputing}, and an Editor of \emph{Journal of Intelligent \& Robotic Systems}.
His current research interests include robotics, distributed parameter systems and intelligent control systems.

\end{document}